\pdfoutput=1
\documentclass[11pt]{article}
\usepackage[final]{emnlp2021}
\usepackage{times}
\usepackage{latexsym}

\usepackage[algo2e,ruled,vlined]{algorithm2e} 
\usepackage{setspace}
\usepackage{verbatim}
\usepackage{xcolor}  
\usepackage{soul}  
\usepackage{bbding}
\usepackage{bm}
\newcommand{\cqrsql}[0]{C{\fontsize{9.2pt}{0}\selectfont QR-SQL}}
\usepackage{multirow}
\usepackage{arydshln}
\usepackage{colortbl}
\usepackage{multirow}
\usepackage{amsmath}
\usepackage{capt-of}
\usepackage{tabularx}
\usepackage[caption=false]{subfig}
\usepackage{epsfig}
\usepackage{amssymb}
\usepackage{amsfonts}
\usepackage{booktabs}
\usepackage{scalerel}
\usepackage[inline]{enumitem}
\usepackage{listings}
\usepackage{varwidth}
\usepackage[export]{adjustbox}
\usepackage{tikz}
\usetikzlibrary{tikzmark}
\usepackage{cleveref}

\usepackage{stmaryrd}
\usepackage{bbm}

\usepackage{algorithm}
\usepackage[noend]{algpseudocode}

\definecolor{deepblue}{rgb}{0,0,0.5}
\definecolor{officeblue}{RGB}{0,102,204}
\definecolor{deepred}{rgb}{0.6,0,0}
\definecolor{deepgreen}{rgb}{0,0.5,0}
\definecolor{mybrickred}{RGB}{182,50,28}

\definecolor{fillcolor}{RGB}{216,217,252}

\algnewcommand\algorithmicrequireb{{\hspace{0.85cm}}}
\algnewcommand\INPTDESCB{\item[\algorithmicrequireb]}

\algnewcommand\algorithmicfuncdesc{\textbf{Function:}}
\algnewcommand\FUNCDESC{\item[\algorithmicfuncdesc]}
\algnewcommand\algorithmicfuncdescb{{\hspace{1.48cm}}}
\algnewcommand\FUNCDESCB{\item[\algorithmicfuncdescb]}
\algnewcommand{\algorithmicgoto}{\textbf{goto}}
\algnewcommand{\Goto}[1]{\algorithmicgoto~\ref{#1}}

\newcommand*\AlgCommentInLine[1]{{\color{deepblue}{$\triangleright$ \textit{#1}}}}

\definecolor{Gray}{gray}{0.92}

\makeatletter
\def\adl@drawiv#1#2#3{%
        \hskip.5\tabcolsep
        \xleaders#3{#2.5\@tempdimb #1{1}#2.5\@tempdimb}%
                #2\z@ plus1fil minus1fil\relax
        \hskip.5\tabcolsep}
\newcommand{\cdashlinelr}[1]{%
  \noalign{\vskip\aboverulesep
           \global\let\@dashdrawstore\adl@draw
           \global\let\adl@draw\adl@drawiv}
  \cdashline{#1}
  \noalign{\global\let\adl@draw\@dashdrawstore
           \vskip\belowrulesep}}
\makeatother

\newcommand{\dashrule}[1][black]{%
  \color{#1}\rule[\dimexpr.2ex-.2pt]{4pt}{.4pt}\xleaders\hbox{\rule{2pt}{0pt}\rule[\dimexpr.2ex-.2pt]{4pt}{.4pt}}\hfill\kern0pt%
}
\newcommand\blfootnote[1]{%
  \begingroup
  \renewcommand\thefootnote{}\footnote{#1}%
  \addtocounter{footnote}{-1}%
  \endgroup
}

\usepackage{microtype}

\title{C{\fontsize{13.1pt}{0}\selectfont QR-SQL}: Conversational Question Reformulation Enhanced Context-Dependent Text-to-SQL Parsers}

\author{
Dongling Xiao$^{1*}$,~~Linzheng Chai$^{2*\dagger}$,~~Qian-Wen Zhang$^1$,~~Zhao Yan$^1$,\\
\textbf{Zhoujun Li}$^2$ and \textbf{Yunbo Cao}$^1$\\
$^1$Tencent Cloud Xiaowei\\
$^{2}$State Key Lab of Software Development Environment, \\Beihang University, Beijing, China\\
$^1${\fontsize{11pt}{0}\selectfont \texttt{\{dlxiao,cowenzhang,zhaoyan,yunbocao\}@tencent.com}} \\$^2${\fontsize{11pt}{0}\selectfont \texttt{\{challenging,lizj\}@buaa.edu.cn}}
}

\begin{document}
\maketitle
\begin{abstract}
Context-dependent text-to-SQL is the task of translating multi-turn questions into database-related SQL queries. Existing methods typically focus on making full use of history context or previously predicted SQL for currently SQL parsing, while neglecting to explicitly comprehend the schema and conversational dependency, such as co-reference, ellipsis and user focus change. In this paper, we propose {\cqrsql}, which uses auxiliary \underline{C}onversational \underline{Q}uestion \underline{R}eformulation (CQR) learning to explicitly exploit schema and decouple contextual dependency for multi-turn SQL parsing. Specifically, we first present a schema enhanced recursive CQR method to produce domain-relevant self-contained questions. Secondly, we train {\cqrsql} models to map the semantics of multi-turn questions and auxiliary self-contained questions into the same latent space through schema grounding consistency task and tree-structured SQL parsing consistency task, which enhances the abilities of SQL parsing by adequately contextual understanding. At the time of writing, our {\cqrsql} achieves new state-of-the-art results on two context-dependent text-to-SQL benchmarks \textsc{SParC}\blfootnote{$^*$ Indicates equal contribution.} and \textsc{CoSQL}\blfootnote{$^\dagger$ The work was done when Linzheng Chai was doing internship at Tencent Cloud Xiaowei.}.
\end{abstract}
\section{Introduction}
\vspace{-0.2cm}
The text-to-SQL task is one of the widely followed branches of semantic parsing, which aims to parse natural language questions with a given database into SQL queries. Previous works \cite{zhongSeq2SQL2017,yu2018spider,wang2020rat} focus on context-independent text-to-SQL task.
\begin{figure}[ht]
	  \centering
	  \setlength{\abovecaptionskip}{4pt}
	  \setlength{\belowcaptionskip}{-15pt}
	   \includegraphics[width=7.7cm]{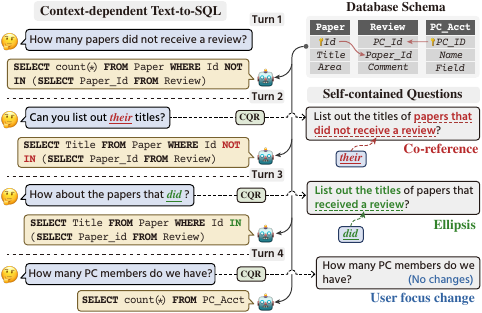}
	   \caption{\textls[-10]{An example of context-dependent Text-to-SQL task demonstrates the phenomenon of co-reference, ellipsis, and user focus changes. The CQR module converts contextual questions to self-contained questions, which can be understood without the context.}}\label{example}
\end{figure}
However, in reality, as users tend to prefer multiple turns interactive queries \cite{iyyer2017search}, the text-to-SQL task based on conversational context is attracting more and more scholarly attention. 
The generalization challenge of the context-dependent text-to-SQL task lies in jointly representing the multi-turn questions and database schema while considering the contextual dependency and schema structure. As shown in Figure \ref{example}, to resolve the contextual dependency, the model should not only understand the \emph{co-reference} and \emph{ellipsis}, but also prevent from  irrelevant information integration when \emph{user focus changes}. Recent studies on two large-scale context-dependent datasets, \textsc{SParC} \cite{yu2019sparc} and \textsc{CoSQL} \cite{yu2020cosql}, also show the difficulty of this problem. To our knowledge, there is a lack of explicit guidance for mainstream text-to-SQL researches dealing with contextual dependency. \vspace{-0.04cm}
\begin{figure*}[t]
\centering 
\setlength{\abovecaptionskip}{4pt}
\setlength{\belowcaptionskip}{-10pt}
\includegraphics[width=16cm]{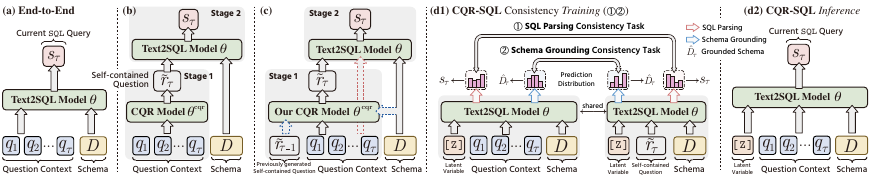}\caption{Schematic of (a) \textbf{End-to-end} and (b) \textbf{Two-stage} pipeline context-dependent text-to-SQL parsing. “Stage 1” in (c) shows the schema enhanced recursive CQR method. (c) A baseline improved on (b) by additionally using question context in “Stage 2”, avoiding model from relying only on potentially incorrect $\tilde{r}_{\tau}$. (d) Our \textbf{C{\fontsize{8.4pt}{0}\selectfont QR-SQL}}.}
\label{fig_preliminary}
\end{figure*}

For context-dependent text-to-SQL, it is common to train a model in an end-to-end manner that simply encoding the concatenation of the multi-turn questions and schema, as shown in Figure \ref{fig_preliminary}(a). To exploit context-dependence information,
\newcite{hui2021dynamic} propose a dynamic relation decay mechanism to model the dynamic relationships between schema and question as conversation proceeds. \newcite{zhang2019editing} and \newcite{hie-sql} leverage previously predicted SQL queries to enhance currently SQL parsing. However, we argue that these end-to-end approaches are inadequate guidance for the contextual dependency phenomenon, though they are competitive in their evaluation of existing context modeling methods. \vspace{-0.025cm}

To help the models achieve adequate understanding of the current user question $q_{\tau}$, conversational question reformulation (CQR) is crucial for multi-turn dialogue systems \cite{pan2019improving,CQA}. 
As far as we know, only few works in con-textual-dependent text-to-SQL, such as \cite{chen2021decoupled}, focus on the value of CQR for modeling question context. \newcite{chen2021decoupled} propose a two-stage pipeline method in which an CQR model first generates a self-contained question $\tilde{r}_{\tau}$, and then a context-independent text-to-SQL parser follows, as shown in Figure \ref{fig_preliminary}(b). But in practice, the limitations of the two-stage pipeline method are in two aspects: 1) the error propagation from the potentially wrong $\tilde{r}_{\tau}$ to the single-turn text-to-SQL parser; 2) the neglect of the relevance between the two stages.  Besides, CQR for text-to-SQL is more challenging than the general CQR tasks \cite{pan2019improving,canard}, since multi-turn questions in text-to-SQL datasets are strictly centered around the underlying database and there are no CQR annotations on existing text-to-SQL datasets. 

Motivated by these observations, we propose {\cqrsql}, which uses auxiliary CQR to achieve adequately contextual understanding, without suffering from the limitations of two-stage methods.
Accordingly, we first introduce an schema enhanced recursive CQR method to product self-contained question data, as in “Stage 1” of Figure \ref{fig_preliminary}(c). The design not only integrates the underlying database schema $D$, but also inherits previous self-contained question $\tilde{r}_{\tau-1}$ to improve the long-range dependency.
Secondly, we propose to train model mapping the self-contained questions and the multi-turn question context into the same latent space through \textbf{schema grounding consistency task} and \textbf{tree-structured SQL parsing consistency task}, as in Figure \ref{fig_preliminary}(d1). 
In this way, to make similar prediction as self-contained question input, models need to pay more attention to the \emph{co-reference} and \emph{ellipsis} when encoding the question context. As shown in Figure \ref{fig_preliminary}(d2), during inference, {\cqrsql} no longer relies on the self-contained questions from CQR models, thus circumventing the error propagation issue of two-stage pipeline methods.

We evaluated {\cqrsql} on \textsc{SparC} and \textsc{CoSQL} datasets, and our main contributions of this work are summarized as follows:\vspace{-0.3cm}
\begin{itemize}[leftmargin=*]
\item We present a schema enhanced recursive CQR mechanism that steadily generates self-contained questions for context-dependent text-to-SQL.\vspace{-0.3cm}
\item 
We propose two novel consistency training tasks to achieve adequate contextual understanding for context-dependent SQL parsing by leveraging auxiliary CQR, which circumvents the limitations of two-stage pipeline approaches.\vspace{-0.3cm}
\item Experimental results show that {\cqrsql} achi-eves state-of-the-art results on context-dependent text-to-SQL benchmarks, \textsc{SparC} and \textsc{CoSQL}, with abilities of adequate context understanding.\vspace{-0.1cm}
\end{itemize}

\section{Proposed Method}
\vspace{-0.2cm}
\label{sec_method}
In this section, we first formally define the context-dependent text-to-SQL task and introduce the backbone network of {\cqrsql}. Afterwards, the technical details of {\cqrsql} are elaborated in two subsections: Schema enhanced recursive CQR and Latent CQR learning for text-to-SQL in context.\vspace{-0.2cm}
\subsection{Preliminary}
\vspace{-0.1cm}
\paragraph{Task Formulation.} In context-dependent text-to-SQL tasks, we are given multi-turn user questions $\bm{q}=\{ q_1,q_2,$ $...,q_n\}$ and the schema $D=\langle T,C \rangle$ of target database which contains a set of tables $T=\{ t_1,t_2,...t_{|T|}\}$ and columns $C_i=\{ c_{i1},c_{i2},...c_{i|C_i|}\}, \forall i\!=\!1,2,...,|T|$ for the $i$-th table $t_i$. Our goal is to generate the target SQL query ${\rm s}_{\tau}$ with the question context $\bm{q}_{\leqslant\tau}$ and schema information $D$ at each question turn $\tau$.\vspace{-0.3cm}
\paragraph{Backbone Network.} {\cqrsql} takes multi-turn questions $\bm{q}$ as input along with the underlying database schema $D$ in the \emph{Encoder-Decoder} framework. For \textbf{encoder}, {\cqrsql} employs the widely used relation-aware Transformer (RAT) encoder \cite{wang2020rat} to jointly represent question and structured schema. For \textbf{decoder}, {\cqrsql} follows the tree-structured LSTM of \newcite{decoder} to predict the grammar rule of SQL abstract syntax tree (AST), column \texttt{id} and table \texttt{id} at each decoding step, indicated as \textsc{ApplyRule}, \textsc{SelectColumn} and \textsc{SelectTable} (See Appendix \ref{appendix_backbone} for detailed descriptions).\vspace{-0.2cm}

\subsection{Schema Enhanced Recursive CQR}
\vspace{-0.1cm}
Due to the scarcity of in-domain CQR annotations for context-dependent text-to-SQL, we adopt self-training with schema enhanced recursive CQR method to collect reliable self-contained questions.\vspace{-0.25cm}

\setlength{\textfloatsep}{0.4cm}
\begin{algorithm}[b]
\caption{Self-training for CQR}
\label{alg:cqr}
\vspace{1pt}
\begin{spacing}{0.8}
{\fontsize{10pt}{0}\selectfont\textbf{Input:} Human-labeled in-domain CQR data $\mathcal{D}_0^{{\rm cqr}}$.}\vspace{-10pt}\\
\end{spacing}
{\fontsize{10pt}{0}\selectfont\textbf{Output:} Full self-contained question data $\mathcal{D}^{\rm cqr}$ for $\mathcal{D}$.}\vspace{-2pt}\\
{\fontsize{10pt}{0}\selectfont$l\leftarrow0$\hfill\AlgCommentInLine{{\fontsize{8pt}{0}\selectfont {\rm Initialize the index of self-training loop.}}}}\\
\While{{\fontsize{10pt}{0}\selectfont $l=0~{\rm or}~|\mathcal{D}_{l-1}^{\rm cqr}|\neq|\mathcal{D}_{l}^{\rm cqr}|$}\vspace{2pt}}
{{\fontsize{10pt}{0}\selectfont 
$\theta^{{\rm cqr}}_{l+1}~\leftarrow\textsc{TrainCQR}(\mathcal{D}_l^{{\rm cqr}})$\vspace{2pt}\\
$\mathcal{D}_{l+1}^{{\rm gen}}\leftarrow\textsc{InferenceCQR}(\mathcal{D},\theta^{{\rm cqr}}_{l+1})$\vspace{2pt}\\
$\mathcal{D}_{l+1}^{{\rm cqr}}\leftarrow\textsc{Union}(\mathcal{D}_{l}^{{\rm cqr}},\textsc{Check}(\theta_{{\rm SQL}}, \mathcal{D}_{l+1}^{{\rm gen}}))$\vspace{-1pt}\\
\begin{spacing}{0.7}
$\!\!$\selectfont \hfill\AlgCommentInLine{{\fontsize{8pt}{0}\selectfont  {\rm \textsc{Check:} Select questions which are self-contained enough for correctly SQL parsing by $\theta_{{\rm SQL}}$ (a pre-trained single-turn text-to-SQL model) in beam search candidates.}}}
\end{spacing}
\vspace{-10pt}
$l\leftarrow l+1$}}
{\fontsize{10pt}{0}\selectfont $\mathcal{D}^{\rm cqr}\leftarrow\textsc{Merge}(\mathcal{D}_{l}^{\rm cqr}, \mathcal{D}_{l}^{\rm gen})$}\hfill\AlgCommentInLine{{\fontsize{8pt}{0}\selectfont {\rm \textsc{Merge:} Replace the self- contained questions in $\mathcal{D}_{l}^{\rm gen}$ with those in $\mathcal{D}_{l}^{\rm cqr}$.}}}\\
\Return{{\fontsize{10pt}{0}\selectfont $\mathcal{D}^{\rm cqr}$}\hfill\AlgCommentInLine{{\fontsize{8pt}{0}\selectfont {\rm Self-contained questions for all interaction turns.}}}}
\end{algorithm}

\paragraph{Schema Integration for CQR.}\!\!Multi-turn ques- tions in text-to-SQL are centered around the underlying database. To generate more domain relevant self-contained question $r_{\tau}$ at each turn $\tau$, we concatenate the question context $\bm{q}_{\leqslant\tau}\!\!$ with schema $D$ as input $\bm{x}_{\tau}\!=\!\{q_1,\texttt{\![\!SEP\!]\!},...,q_{\tau},\texttt{\![\!SEP\!]\!},t_1,c_{11},$ $c_{12},...,\texttt{\![\!SEP\!]\!},t_2,c_{21},c_{22},...\}$ for CQR learning.\vspace{-0.2cm}

\paragraph{Recursive Generation for CQR.}
Inspired by \newcite{zhang2019editing} and \newcite{istsql}, who verify that integration of previously predicted SQL facilitates modeling long interactions turns, we propose a recursive generation mechanism to recursively inherit context information from previously generated self-contained questions $\tilde{r}_{\tau-1}$ for long-range dependence, as shown in the stage 1 of Figure \ref{fig_preliminary}(c). Our CQR at each turn $\tau$ is optimized as:\begin{equation}
\setlength{\abovedisplayskip}{5pt}
\setlength{\belowdisplayskip}{5pt}
\resizebox{.73\linewidth}{!}{$
    \displaystyle
    \mathcal{L}^{{\rm cqr}}_{\tau}\!=\!-{\rm log}\mathcal{P}(r_{\tau}|\{ \tilde{r}_{\tau-1},\texttt{\![\!SEP\!]\!},\bm{x}_{\tau}\}).$}
\end{equation}

During training, other than using the labeled self-contained questions $r_{\tau-1}$ as $\tilde{r}_{\tau-1}$, we sampled $\tilde{r}_{\tau-1}$ from a pre-trained CQR model to reduce discrepancies between training and inference.\vspace{-0.2cm}

\paragraph{Self-training for CQR.}
\label{sec2.2}
\newcite{chen2021decoupled} indicate that models trained with general CQR datasets work poor on the in-domain data from \textsc{\textsc{CoSQL}} and \textsc{SParC}. Besides the annotated in-domain self-contained question data is scarce for all context-dependent text-to-SQL tasks.

We conduct a self-training approach with a pre-trained single-turn text-to-SQL model $\theta_{{\rm SQL}}$ to collect full self-contained question data $\mathcal{D}^{\rm cqr}$ for text-to-SQL datasets $\mathcal{D}$, as show in Algorithm \ref{alg:cqr}.
\subsection{C{\fontsize{10pt}{0}\selectfont QR-SQL}~: Latent CQR Learning for Text-to-SQL Parsing in Context} 
With the self-contained questions $\mathcal{D}^{\rm cqr}$ in \S\ref{sec2.2}, during training, we introduce \cqrsql, which uses a latent variable $\!\texttt{[Z]}\!$ to map the semantics of question context and self-contained question into the same latent space with two consistency tasks (schema grounding and SQL parsing), helping models achieve adequately contextual understanding for enhanced SQL parsing during inference. 

\begin{figure*}
	\centering
	\setlength{\abovecaptionskip}{3pt}
    \setlength{\belowcaptionskip}{-7pt}
	\includegraphics[width=16cm]{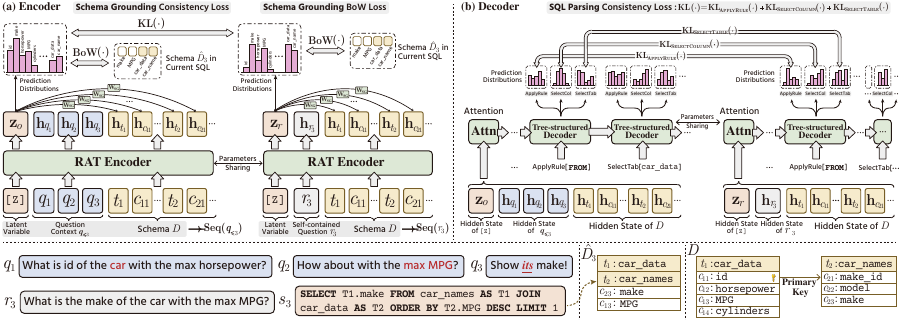}
	\caption{
	Illustration of the \emph{training stage} for \cqrsql. (a) Schema grounding task with bag-of-word (BoW) loss and consistency loss. (b) Tree-structured SQL parsing consistency loss at each decoding step. $\texttt{\![Z]\!}$ denotes the special symbol of latent variable. RAT Encoder is the relation-aware transformer encoder \protect\cite{wang2020rat} to jointly represent natural language and structured schema. Tree-structured Decoder is the tree-structured LSTM of \newcite{decoder} to predict SQL AST rules, Table \texttt{id} and Column \texttt{id} at each decoding step.}
	\label{fig_overview}
\end{figure*}

As shown in Figure \ref{fig_overview}(a), during training, we input ${\rm Seq}(q)\!=\!\{\texttt{[Z]},q,\texttt{[\!SEP\!]},D\}$ to \cqrsql, where $q$ can be the question context $\bm{q}_{\leqslant\tau}$ or self-contained questions $r_{\tau}$.\vspace{-0.15cm}

\paragraph{\textls[-25]{Schema Grounding Consistency Task.}} Ground- ing tables and columns into question context requires adequately understanding the \emph{co-reference} and \emph{ellipsis} in multi-turn questions. Thus we propose using the hidden state $\bm{{\rm z}}$ of latent variable to predict the tables and columns appear in current target SQL query $s_{\tau}$ with bag-of-word (BoW) loss \cite{bow}, and then enforcing models to make consistent predictions with question context input and self-contained question input, as shown in Figure \ref{fig_overview}(a). The BoW loss of \underline{S}chema \underline{G}round-ing task $\mathcal{L}^{{\rm SG}_{{\rm BoW}}}$ at each turn $\tau$ are formulated as:\begin{equation}
\setlength{\abovedisplayskip}{6pt}
\setlength{\belowdisplayskip}{0pt}
\resizebox{.72\linewidth}{!}{$
    \displaystyle
    \mathcal{L}^{{\rm SG}_{{\rm BoW}}}_{\tau}={\rm BoW}(\bm{q}_{\leqslant\tau})+{\rm BoW}(r_{\tau}).
$}
\end{equation}
\begin{equation}
\setlength{\abovedisplayskip}{0pt}
\setlength{\belowdisplayskip}{0pt}
\resizebox{.85\linewidth}{!}{$
    \displaystyle
    \begin{aligned}
    {\rm BoW}(q)\!&=\!-{\rm log}\mathcal{P}(\hat{D}_{\tau}|{\rm Seq}(q))\\ 
    &=\!-\!\!\!\!\!\sum_{~~\hat{d}\in\hat{D}_{\tau}}\!\!\!\!{\rm log}\frac{e^{f\!_{\hat{d}}({\rm RAT}({\rm Seq}(q))_0)}}{\sum_{d\in D} e^{f_d({\rm RAT}({\rm Seq}(q))_0)}}.
\end{aligned}
$}
\end{equation}

where $\hat{D}_{\tau}$ refers to the schema appeared in current SQL query $s_{\tau}$, $D$ indicates the full schema of target database. $\mathcal{P}(\hat{D}_{\tau}|\cdot)$ represents the schema prediction probability distributions at turn $\tau$. The function $f_d(\bm{{\rm z}})=\bm{{\rm h}}_d\bm{{\rm W}}_{\rm \!SG}\bm{{\rm z}}^{\!\top}$, $\bm{{\rm h}}_d$ denotes the final hidden states of schema $d$ for ${\rm RAT}$ encoder. $\bm{{\rm z}}_o\!=\!{\rm RAT}({\rm Seq}(\bm{q}_{\leqslant\tau}))_0$ and $\bm{{\rm z}}_r\!=$ ${\rm RAT}({\rm Seq}(r_{\tau}))_0$  indicate the final hidden state of the latent variables associated with question context $\bm{q}_{\leqslant\tau}$ and self-contained question $r_{\tau}$ respectively. The \underline{S}chema \underline{G}rounding consistency loss $\mathcal{L}^{{\rm SG_{KL}}}$ is defined as:
\begin{equation}
\setlength{\abovedisplayskip}{9pt}
\setlength{\belowdisplayskip}{9pt}
\resizebox{.89\linewidth}{!}{$
    \displaystyle
    \begin{aligned}
    \hspace{-0.5em}
    \mathcal{L}^{{\rm SG_{KL}}}_{\tau}\!&=\!{\rm KL}\!\left(\!\mathcal{P}(\hat{D}_{\tau}|{\rm Seq}(\bm{q}_{\leqslant\tau}))\!\parallel\!\mathcal{P}(\hat{D}_{\tau}|{\rm Seq}(r_{\tau}))\!\right)  \\
    \!&+{\rm KL}\!\left(\!\mathcal{P}(\hat{D}_{\tau}|{\rm Seq}(r_{\tau}))\!\parallel\!\mathcal{P}(\hat{D}_{\tau}|{\rm Seq}(\bm{q}_{\leqslant\tau}))\!\right)\!.
    \hspace{-2em}
    \end{aligned}
$}
\end{equation}
where ${\rm KL}(\cdot)$ refers to the Kullback–Leibler divergence between two distributions.\vspace{-0.2cm}
\paragraph{SQL Parsing Consistency Task.}
Furthermore, to encourage model pay more attention to the SQL logic involving co-reference and ellipsis, we introduce to enforce the model to obtain the consistency prediction of SQL parsing with question contexts and self-contained questions as inputs, at each decoding step. The SQL parsing loss $\mathcal{L}^{{\rm SP}}$ and the \underline{S}QL \underline{P}arsing consistency loss $\mathcal{L}^{{\rm SP_{KL}}}$, at each turn $\tau$, can be represented as:
\begin{equation}
\setlength{\abovedisplayskip}{8pt}
\setlength{\belowdisplayskip}{0pt}
\resizebox{.89\linewidth}{!}{$
    \displaystyle
    \hspace{-0.5em}
    \mathcal{L}^{{\rm SP}}_{\tau}\!\!=\!-{\rm log}\mathcal{P}(s_{\tau}|{\rm Seq}(\bm{q}_{\leqslant\tau}))\!-\!{\rm log}\mathcal{P}(s_{\tau}|{\rm Seq}(r_{\tau})).
    \hspace{-0.5em}$}
\end{equation}
\begin{equation}
\setlength{\abovedisplayskip}{0pt}
\setlength{\belowdisplayskip}{8pt}
\resizebox{.89\linewidth}{!}{$
    \displaystyle
    \begin{aligned}
    \hspace{-0.5em}
    \mathcal{L}^{{\rm SP_{KL}}}_{\tau}\!&=\!{\rm KL}\!\left(\mathcal{P}(s_{\tau}|{\rm Seq}(\bm{q}_{\leqslant\tau}))\!\parallel\!\mathcal{P}(s_{\tau}|{\rm Seq}(r_{\tau}))\right)  \\
    \!&+{\rm KL}
    \!\left(\mathcal{P}(s_{\tau}|{\rm Seq}(r_{\tau}))\!\parallel\!\mathcal{P}(s_{\tau}|{\rm Seq}(\bm{q}_{\leqslant\tau}))\right)\!.
    \hspace{-2em}
    \end{aligned}
$}
\end{equation}

In this work, we follow the tree-structured decoder of \newcite{decoder}, which generates SQL queries as an abstract syntax tree (AST), and conduct three main predictions at each decoding step, including \textsc{ApplyRule}, \textsc{SelectColumn} and \textsc{SelectTable}. We calculate the SQL parsing consistency loss by accumulating all KL divergences of above three predictions as ${\rm KL}(\cdot)\!=\!{\rm KL}_{\textsc{ApplyRule}}(\cdot)\!+\!{\rm KL}_{\textsc{SelectColumn}}(\cdot)\!+\!{\rm KL}_{\textsc{SelectTable}}(\cdot)$ at all decoding steps, as shown in Figure \ref{fig_overview}(b) and further described in Appendix~\ref{appendix_sp}. 

Finally we calculate the total training loss $ \mathcal{L}_{\tau}$ at each question turn $\tau$ for our context-dependent text-to-SQL model {\cqrsql} as:
\begin{equation}
\setlength{\abovedisplayskip}{10pt}
\setlength{\belowdisplayskip}{5pt}
\resizebox{.88\linewidth}{!}{$
    \displaystyle
    \mathcal{L}_{\tau}=\mathcal{L}^{\rm SP}_{\tau}\!+\!\lambda_1\mathcal{L}^{\rm SG_{{}\rm BoW}}_{\tau}\!\!+\underbrace{\!\lambda_2\!\left(\mathcal{L}^{\rm SP_{KL}}_{\tau}\!+\!\mathcal{L}^{\rm SG_{KL}}_{\tau}\right)\!}_{{\rm Consistency~Loss}}.$}
\end{equation}
where $\lambda_1$ and $\lambda_2$ are weights for the schema grounding BoW loss and the consistency loss respectively.\vspace{-0.1cm}

\paragraph{\textls[-25]{{\cqrsql} Inference.}\!\!\!\!} Since {\cqrsql} has learned to adequately understand the context dependency in question context $\bm{q}_{\leqslant\tau}$ by distilling representations from self-contained question in two consistency tasks, {\cqrsql} no longer relies on self-contained questions and only considers ${\rm Seq}(\bm{q}_{\leqslant\tau})$ as inputs, as shown in Figure \ref{fig_preliminary}(d2), thus circumventing the error propagation in two-stage pipeline methods. \vspace{-0.1cm}

\begin{table*}[t]
\centering
\setlength{\abovecaptionskip}{4pt}
\setlength{\belowcaptionskip}{-8pt}
\addtolength{\tabcolsep}{-0.8mm}
\resizebox{0.989\textwidth}{!}{
\begin{tabular}{l|cccccccc}
\toprule[1.0pt]
 \multirow{2}{*}{\textbf{Models}~($\downarrow$)~/~\textbf{Datasets}~($\rightarrow$)}&\multicolumn{2}{c}{\textbf{\textsc{SParC}$_{\bm{{\rm Dev}}}$}}& \multicolumn{2}{c}{\textbf{\textsc{SParC}$_{\bm{{\rm Test}}}$}}&\multicolumn{2}{c}{\textbf{\textsc{CoSQL}}$_{\bm{{\rm Dev}}}$}&\multicolumn{2}{c}{\textbf{\textsc{CoSQL}}$_{\bm{{\rm Test}}}$} \\
&${\rm QM_{{\scriptscriptstyle (\%)}}}$&${\rm IM _{{\scriptscriptstyle (\%)}}}$&${\rm QM _{{\scriptscriptstyle (\%)}}}$&${\rm IM _{{\scriptscriptstyle (\%)}}}$&${\rm QM _{{\scriptscriptstyle (\%)}}}$&${\rm IM _{{\scriptscriptstyle (\%)}}}$&${\rm QM _{{\scriptscriptstyle (\%)}}}$&${\rm IM _{{\scriptscriptstyle (\%)}}}$ \\
\midrule[0.3pt]
 GAZP~+~BERT \cite{zhong-etal-2020-grounded}&48.9&29.7&45.9&23.5&42.0&12.3&39.7&12.8\\
 IGSQL~+~BERT \cite{cai-wan-2020-igsql}&50.7&32.5&51.2&29.5&44.1&15.8&42.5&15.0\\
 R$^2$SQL~+~BERT \cite{hui2021dynamic}&54.1&35.2&55.8&30.8&45.7&19.5&46.8&17.0\\
 RAT-SQL~+~BERT \cite{yu2021score}&56.8&33.4&-&-&48.4&19.1&-&-\\
 DELTA~+~BERT$^{\heartsuit}$ \cite{chen2021decoupled}&58.6&35.6&59.9&31.8&51.7&21.5&50.8&19.7\\
 \textbf{C{\fontsize{9.5pt}{0}\selectfont QR-SQL}}~+~BERT (Ours)&\textbf{62.5}&\textbf{42.4}&-&-&\textbf{53.5}&\textbf{24.6}&-&-\\
\cdashlinelr{1-9}
 RAT-SQL~+~\textsc{SCoRe}$^{\diamondsuit}$ \cite{yu2021score}&62.2&42.5&62.4&38.1&52.1&22.0&51.6&21.2\\
 RAT-SQL~+~TC~+~\textsc{Gap}$^{\diamondsuit}$ \cite{rat-sql-tc}&64.1&44.1&65.7&43.2&-&-&-&-\\
 \textsc{Picard}~+~T5-3B \cite{scholak2021picard}&-&-&-&-&56.9&24.2&54.6&23.7\\
 HIE-SQL~+~\textsc{Grappa}$^{\diamondsuit}$ \cite{hie-sql}~~&64.7&45.0&64.6&42.9&56.4&28.7&53.9&24.6\\
 \textsc{UnifiedSKG}~+~T5-3B \cite{unifiedskg}&61.5& 41.9&-&-& 54.1&22.8&-&-\\

 RASAT~+~T5-3B \cite{rasat}&66.7&47.2&-&-&\textbf{58.8}&26.3&-&-\\
\midrule[0.3pt]
 \textbf{C{\fontsize{9.5pt}{0}\selectfont QR-SQL}}~+~\textsc{Electra}~(Ours)&67.8&48.1&67.3&43.9&58.4&29.4&-&-\\
 \textbf{C{\fontsize{9.5pt}{0}\selectfont QR-SQL}}~+~\textsc{Coco-LM}~(Ours)&\textbf{68.0}&\textbf{48.8}&\textbf{68.2}&\textbf{44.4}&58.5&\textbf{31.1}&\textbf{58.3}&\textbf{27.4}\\
\bottomrule[1.0pt]
\end{tabular}}
\caption{Performances on the development and test set of \textsc{SParC} and \textsc{CoSQL}. ``${\bm{{\rm QM}}}$" and ``$\bm{{{\rm IM}}}$" indicate the exact match accuracy over all questions and all interaction respectively. 
The models with $\!^\diamondsuit\!$ mark employ task adaptive pre-trained language models. Models with $\!^\heartsuit\!$ mark use the general two-stage pipeline approach in Figure \ref{fig_preliminary}(b). The ``-" results of {\cqrsql} are awaiting evaluation due to the submission interval of the leaderboard.} 
\label{tab_dev_result}
\end{table*}

\section{Experiments}
\vspace{-0.15cm}
In this section, we conduct several experiments to assess the performance of proposed methods in \S\ref{sec_method}.\vspace{-0.2cm}
\subsection{Experimental Setup}
\vspace{-0.1cm}
\paragraph{CQR Learning.\!\!\!}We adopt the Transformer-based encoder-decoder architecture based on the pre-trained ProphetNet \cite{prophetnet} as the initial CQR model. Since there is no question reformulation annotations in \textsc{SParC} and \textsc{CoSQL}, we annotate $3034$ and $1527$ user questions as the initial in-domain supervised CQR data $\mathcal{D}^{\rm cqr}_0$ for \textsc{SParC} and \textsc{CoSQL} respectively. Before self-training, we pre-train a single-turn text-to-SQL model $\theta_{{\rm SQL}}$ based on RAT-SQL \cite{wang2020rat} architecture and \textsc{Electra} \cite{clark2020electra} language model for checking whether a generated question is self-contained enough for correctly SQL parsing. During self-training in \S\ref{sec2.2}, we conduct $3$ training loops $\{ \theta^{{\rm cqr}}_1$, $\theta^{{\rm cqr}}_2$, $\theta^{{\rm cqr}}_3 \}$  and obtain $4441$ and $1973$ supervised CQR data for \textsc{SParC} and \textsc{CoSQL} respectively. Finally, we use the CQR model $\theta^{{\rm cqr}}_3$ in the last training loop to produce the self-contained questions for all interaction turns.\vspace{-0.25cm}

\begin{table}[t]
\centering
\setlength{\abovecaptionskip}{4pt}
\setlength{\belowcaptionskip}{-5pt}
\addtolength{\tabcolsep}{-1mm}
\resizebox{0.485\textwidth}{!}{
\begin{tabular}{l|cccc}  
\toprule[1.0pt]
\multirow{2}{*}{\textbf{Dataset}} & \#~Num of & \multirow{2}{*}{\#Train/\#Dev/\#Test}&\#Average & System  \\
&Interactions&&Turn&Response   \\
\midrule[0.3pt]
\textsc{SParC}&4,298 &3,034 / 422 / 842 &3.0&\XSolidBrush\\
\textsc{CoSQL}&3,007 &2,164 / 293 / 551 &5.2&\Checkmark\\
\bottomrule[1.0pt]	
\end{tabular}}
\caption{Detailed statistics for \textsc{SParC} and \textsc{CoSQL}.}
\label{tab_statistic_information}
\end{table}

\paragraph{C{\fontsize{9.3pt}{0}\selectfont QR-SQL} Training.\!\!\!} We conduct experiments on two context-dependent text-to-SQL datasets \textsc{SParC} and \textsc{CoSQL}, the statistic information of them are depicted in Table \ref{tab_statistic_information}. Following \cite{cao2021lge}, we employ RAT-SQL \cite{wang2020rat} architecture and pre-trained \textsc{Electra} \cite{clark2020electra} for all text-to-SQL experiments in this paper. In the training of {\cqrsql}, we set hyperparameters $\lambda_1=0.1$ and $\lambda_2=3.0$ for \textsc{SParC}, $\lambda_2=1.0$ for \textsc{CoSQL} (See Appendix~\ref{appendix_weight} for details), learning rate as $5e$-$5$, batch size of $32$. During inference, we set the beam size to $5$ for SQL parsing.\vspace{-0.2cm}

\subsection{Experimental Results}
\vspace{-0.1cm}
As shown in Table \ref{tab_dev_result}, {\cqrsql} achieves state-of-the-art results cross all settings at the time of writing. With general PLM BERT, {\cqrsql} surpasses all previous methods, including the two-stage method DELTA \cite{chen2021decoupled} which also uses additional text-to-SQL data from Spider. Beside, most of recent advanced methods tend to incorporates more task-adaptive data (\texttt{text-table} pairs and synthesized \texttt{text-sql} pairs), tailored pre-training tasks (column prediction and turn switch prediction) and super-large PLM T5-3B~\cite{t5} into training. For this setting, we use general PLM \textsc{Electra} for all text-to-SQL experiments following~\newcite{cao2021lge}, and further employ a more compatible\footnotemark\!~PLM \textsc{Coco-LM}~\cite{coco} for comparison. {\cqrsql} significantly outperforms \textsc{SCoRe} \cite{yu2021score}, RAT-SQL+TC \cite{rat-sql-tc} and recent HIE-SQL~\cite{hie-sql} which use task-adaptive pre-trained models. Note that HIE-SQL employs two task-adaptive PLMs for encoding \texttt{text-schema} pairs and previous SQL queries respectively. Compared with methods based on super-large T5-3B model (especially RASAT~\cite{rasat} which integrates co-reference relations and constrained decoding into T5-3B), {\cqrsql} can also achieve significant improvements.  
 
\footnotetext[1]{\textsc{Coco-LM} is pre-trained on sequence contrastive
learning with a dual-encoder architecture~\cite{sbert}, which is compatible for our CQR consistency tasks with dual-encoder for multi-turn $\bm{q}_{\leqslant\tau}$ and self-contained $r_{\tau}$.}

\begin{figure}[t]
\centering 
\setlength{\abovecaptionskip}{3pt}
\setlength{\belowcaptionskip}{2pt}
\includegraphics[width=7.7cm]{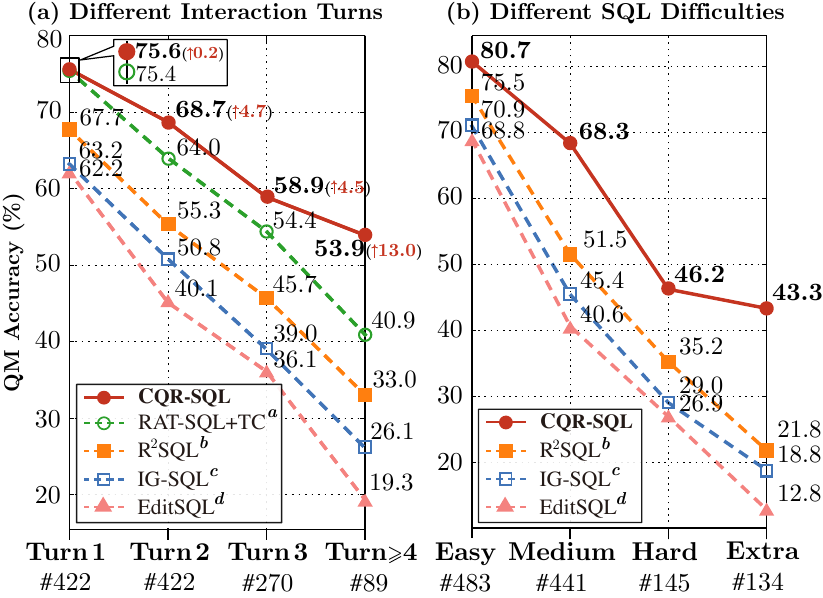}\caption{Detailed question match ($\bm{{\rm QM}}$) accuracy results in different \textbf{interaction turns} and \textbf{goal difficulties} on the dev set of \textsc{SParC} dataset. \# ${\rm Number}$ denotes the number of questions. Detailed results of $^{a\;}$\protect\cite{rat-sql-tc},$^{b\;}$\protect\cite{hui2021dynamic},$^{c\;}$\protect\cite{cai-wan-2020-igsql} and $^{d\;}$\protect\cite{zhang2019editing} are from the original paper.}
\label{fig_detail}
\end{figure}

To verify the advantages of {\cqrsql} on adequately contextual understanding, we further compare the performances on different interaction turns of \textsc{SParC}, as shown in Figure \ref{fig_detail}(a). We observe that it is more difficult for SQL parsing in longer interaction turns due to the long-range dependency problem, while {\cqrsql} achieves more significant improvement as the interaction turn increases. Moreover, in Figure \ref{fig_detail}(b), we further compare the performances on varying difficulty levels of target SQL queries, {\cqrsql} consistently outperforms previous works on all difficulty levels, especially on the “$\bm{{\rm Extra~Hard}}$” level whose target SQL queries are most complex and usually contain nesting SQL structures~\cite{yu2018spider}.\vspace{-0.2cm}

\setlength{\textfloatsep}{0.3cm}
\begin{table}[b]
\centering
\setlength{\abovecaptionskip}{3pt}
\setlength{\belowcaptionskip}{-10pt}
\addtolength{\tabcolsep}{-0.5mm}
\resizebox{0.485\textwidth}{!}{
\begin{tabular}{l|ccc}  
\toprule[1.0pt]
\hspace{-0.2em}\textbf{Models~/~Task}&\textbf{\textsc{SParC}}$_{\bm{{\rm CQR}}}\!\!$&\textbf{\textsc{CoSQL}}$_{\bm{{\rm CQR}}}\!\!$&\textbf{\textsc{CoSQL}}\\
\hspace{-0.2em}\#~Train / \#~Dev&$3,\!034$ / $422$&$1,\!527$ / $154$&${\rm QM}$~/~${\rm IM}$\\
\midrule[0.3pt]
\hspace{-0.2em}\textbf{Our CQR}&55.75 / \textbf{73.80}&\textbf{59.87} / \textbf{78.82}&\textbf{57.8} / \textbf{27.7}\\
\hspace{-0.2em}$~-{\rm RG}$&\textbf{55.83} / 73.72&59.41 / 78.24&57.3 / 27.3\\
\hspace{-0.2em}$~-{\rm RG}-{\rm SE}$&54.78 / 72.87&58.93 / 77.64&56.4 / 26.6\\
\cdashlinelr{1-4}
\hspace{-0.2em}\textsc{Canard}$_{{\rm T5~3B}}$&25.57 / 49.52&39.04 / 57.08& 53.6 / 23.9\\
\bottomrule[1.0pt]
\end{tabular}
}
\caption{Comparisons on the ${\rm BLEU}$ / ${\rm Rouge}$-${\rm L}$ scores between schema enhanced (${\rm SE}$) approach and recursive generation (${\rm RG}$) for CQR. Models are trained and evaluated on the initially annotated data $\mathcal{D}_0^{{\rm cqr}}$. We observe that the ${\rm BLEU}$ / ${\rm Rouge}$-${\rm L}$ scores of \textsc{CoSQL}$_{\rm CQR}$ are much higher than those of \textsc{SParC}, because \textsc{CoSQL} has much more user focus change questions that without co-references and ellipsis \protect\cite{yu2020cosql}.}
\label{tab_cqr_ablation}
\end{table}

\subsection{Ablation Study}
\vspace{-0.1cm}
Regarding the CQR task, as shown in Table \ref{tab_cqr_ablation}, recursive generation (${\rm RG}$) achieves $0.46\%$ BLEU score gains on the CQR task for \textsc{CoSQL} dataset which has much longer interaction turns than \textsc{SParC} as shown in Table \ref{tab_statistic_information}, while ${\rm RG}$ fails to significantly improve the performance for \textsc{SParC}$_{\rm CQR}$. This indicates ${\rm RG}$ can improve CQR performance for longer contextual dependency. While further removing the schema enhanced ${\rm SE}$ method, performances decrease by roughly $1\%$ and $0.5\%$ on \textsc{SParC}$_{\rm CQR}$ and \textsc{CoSQL}$_{\rm CQR}$ respectively, which verifies the effectiveness of schema integration in CQR for text-to-SQL datasets. We additionally evaluate the performances without any in-domain CQR annotations (fine-tune T5-3B on general CQR dataset \textsc{Canard}~\cite{canard}), and observe that performances on CQR tasks are more significantly reduced than on \textsc{CoSQL}, verifying the effect of in-domain data and the robustness of {\cqrsql} against noised CQR information.

\begin{table}[t]
\centering
\setlength{\abovecaptionskip}{3pt}
\setlength{\belowcaptionskip}{2pt}
\addtolength{\tabcolsep}{-0.8mm}
\resizebox{0.480\textwidth}{!}{
\begin{tabular}{ll|cccc} 
\toprule[1.0pt]
\multirow{2}{*}{\!\!\texttt{[\#]\!\!}}&\textbf{Datasets}~($\rightarrow$)&\multicolumn{2}{c}{\textbf{\textsc{SParC}}$_{\bm{{\rm Dev}}}$}&\multicolumn{2}{c}{\textbf{\textsc{CoSQL}}$_{\bm{{\rm Dev}}}$}\\
&\textbf{Models}~($\downarrow$)&${\rm QM_{{\scriptscriptstyle (\%)}}}$&${\rm IM_{{\scriptscriptstyle (\%)}}}$&${\rm QM_{{\scriptscriptstyle (\%)}}}$&${\rm IM_{{\scriptscriptstyle (\%)}}}$\\
\midrule[0.3pt]


\texttt{\!\![1]\!\!}&\textbf{C{\fontsize{9.5pt}{0}\selectfont QR-SQL}}&\textbf{67.8}&\textbf{48.1}&\textbf{58.4}&\textbf{29.4}\\
\texttt{\!\![2]\!\!}&~$-{\rm SG}$&66.3&47.4&57.0&27.0\\
\texttt{\!\![3]\!\!}&~$-{\rm SP_{KL}}$&65.6&46.9&57.3&25.6\\
\texttt{\!\![4]\!\!}&~$-{\rm SP_{KL}-\!SG}$ &64.9&46.5&56.6&23.9\\
\texttt{\!\![5]\!\!}&~$-{\rm SP_{KL}-\!SG_{KL}}$&64.7&45.7&56.1&23.6\vspace{-2pt}\\
\specialrule{1pt}{3pt}{0pt}
\rowcolor{Gray}
\multicolumn{6}{l}{\textbf{Exploratory Study} on the Integration of CQR} \vspace{-2pt}\\
\specialrule{.5pt}{2pt}{3pt}
\texttt{\!\![6]\!\!}&CQR$_{\rm Augment}$&64.5&45.7&54.7&24.2\\
\texttt{\!\![7]\!\!}&CQR$_{\rm Two~Stage^{\star}}$&65.8&46.7&56.8&24.6\\
\texttt{\!\![8]\!\!}&CQR$_{\rm Two~Stage}$&62.5&43.1&54.8&23.6\\
\texttt{\!\![9]\!\!}&CQR$_{\rm Multi~Task}$&64.9&45.0&56.2&25.3\\
\bottomrule[1.0pt]
\end{tabular}
}
\caption{Ablation studies for C{\fontsize{8.4pt}{0}\selectfont QR-SQL} and its variants. ${\rm SG}$ denotes the \underline{S}chema \underline{G}rounding task (including ${\rm BoW}$ loss and consistency loss ${\rm SG_{\rm KL}}$), and ${\rm SP_{KL}}$ denotes the \underline{S}QL \underline{P}arsing consistency task.}
\label{tab_cqr_sql_ablation}
\end{table}

\begin{figure*}[t]
\centering 
\setlength{\abovecaptionskip}{3pt}
\setlength{\belowcaptionskip}{-8pt}
\includegraphics[width=16cm]{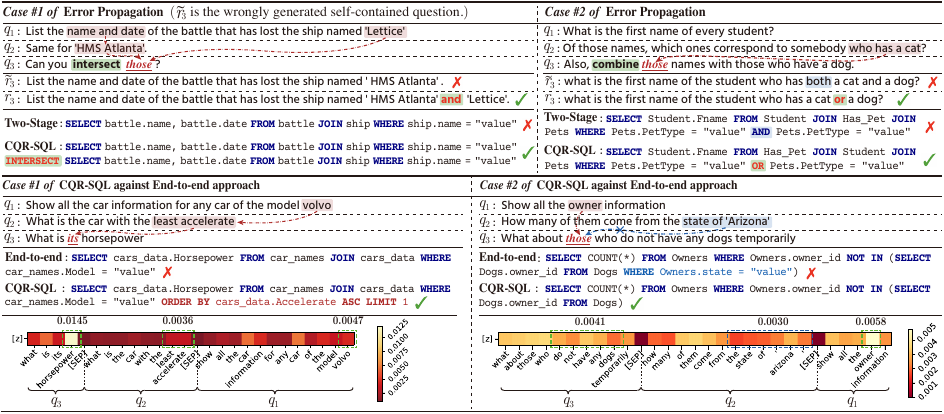}
\caption{Case studies on \textsc{SParC} dataset. Upper block shows the cases of error propagation with incorrectly generated self-contained questions $\tilde{r}_{\tau}$ for \textbf{Two-Stage} pipeline methods (as in Figure \ref{fig_preliminary}(c) or \texttt{[7]} in Table \ref{tab_cqr_ablation}). Cases in the lower block show that \textbf{End-to-End} method (as in Figure \ref{fig_preliminary}(a) or \texttt{\![4]\!} in Table \ref{tab_cqr_sql_ablation}) fails to resolve the conversational dependency. Besides, the heat maps represent the visualization of attention scores of the latent variable \texttt{[Z]} on the question context part. Number on each dotted box is the average attention scores of that box.}
\label{fig_case_study}
\end{figure*}

\begin{figure}[h]
\centering 
\setlength{\abovecaptionskip}{3pt}
\setlength{\belowcaptionskip}{-8pt}
\includegraphics[width=7.5cm]{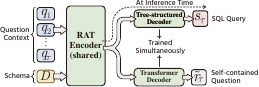}
\caption{Schematic of CQR$_{{\rm Multi~Task}}$ variant.}
\label{fig_multitask}
\end{figure}

In Table \ref{tab_cqr_sql_ablation}, we investigate the contribution of each designed choice of proposed {\cqrsql}.\vspace{-5pt}
\paragraph{{\rm $\texttt{[2]}\!:\!-{\rm SG}.\!\!\!\!$}} If removing schema grounding task (including ${\rm BoW}$ loss and consistency loss), all metrics on \textsc{SParC} and \textsc{CoSQL} drops by $0.7\%$-$2.4\%$.\vspace{-4pt}
\paragraph{{\rm $\texttt{[3]}\!:\!-{\rm SP_{KL}}.\!\!\!\!$}} After removing SQL parsing consistency loss, all metrics drops by $1.1\%$-$3.8\%$. \vspace{-5pt}
\paragraph{{\rm $\texttt{[4]}\!\!:\!\!-{\rm SP_{KL}-\! SG}.$}} If removing both schema grounding task (${\rm BoW}$ loss and consistency loss) and SQL parsing consistency loss, {\cqrsql} degenerates to a variant trained in the general “\textbf{End-to-End}” manner as shown in Figure \ref{fig_preliminary}(a), and its performances decrease by $1.7\%$-$5.5\%$.\vspace{-5pt} 
\paragraph{{\rm \texttt{[5]}: $-{\rm SP_{KL}-\!SG_{KL}}.$}} To study the impact of schema grounding ${\rm BoW}$ loss, we train End-to-End variants \texttt{[4]} with ${\rm {BoW}}$ loss and find that merely integrating ${\rm {BoW}}$ loss slightly degrades the performances, because grounding schema in context is much more difficult than in single questions. While combining ${\rm {BoW}}$ with consistency loss to distill contextual knowledge from self-contained questions can improve the performances (\texttt{\![4]\!}$\rightarrow$\texttt{\![3]\!}).\vspace{-4pt} 
\paragraph{{\rm \texttt{[6]}:~CQR$_{{\rm Augment}}.\!\!\!$}} We directly augment the full self-contained question data $\mathcal{D}^{\rm cqr}$ to \texttt{[4]} as single-turn text-to-SQL data augmentation. We find CQR$_{{\rm Augment}}$ degrades the performances because models are trained more on single-turn text-to-SQL, while weaken the abilities of contextual understanding for multi-turn text-to-SQL.\vspace{-4pt} 
\paragraph{{\rm \texttt{[7]}:~CQR$_{{\rm Two~Stage^{\star}}}.\!\!\!$}} A variant trained in the improved “\textbf{Two-Stage}” manner as in Figure \ref{fig_preliminary}(c). It slightly outperforms the End-to-End variant \texttt{\![4]\!}.\vspace{-4pt} 
\paragraph{{\rm \texttt{[8]}:~CQR$_{{\rm Two~Stage}}.\!\!\!$}} A “Two-Stage” variant without additionally integrating question context into “Stage 2” as variant \texttt{\![7]\!}. The performances significantly decline for the error propagation issue.\vspace{-4pt} 
\paragraph{{\rm \texttt{[9]}:~CQR$_{{\rm Multi~Task}}.\!\!\!$}} A variant jointly trained on CQR task and text-to-SQL task with a shared RAT encoder and two task-specified decoders, as shown in Figure \ref{fig_multitask}. We observe that CQR$_{{\rm Multi~Task}}$ slightly decreases the performances compared with End-to-End variant \texttt{\![4]\!} for the optimization gap between the \texttt{text-to-text} optimization and structured \texttt{text-to-SQL} optimization.\vspace{-0.1cm}

\setlength{\textfloatsep}{0.0cm}

Above results and analyses demonstrate the advantages of {\cqrsql} on leveraging self-contained questions to enhance the abilities of adequately context understanding for contextual text-to-SQL parsing, meanwhile circumventing the error propagation of Two-Stage pipeline methods.\vspace{-0.1cm}

\subsection{Case Study}
\vspace{-0.1cm}
As shown in the upper block of Figure \ref{fig_case_study}, we compare the SQL queries by {\cqrsql} with those by the Two-Stage baseline model in Figure \ref{fig_preliminary}(c) or \texttt{[7]} in Table \ref{tab_cqr_ablation}, to show the error propagation phenomenon between CQR stage and text-to-SQL stage. The generated self-contained questions $\tilde{r}_{\tau}$ are incorrect, leading to wrong predictions of SQL queries. Specifically, In the first case, CQR model fails to understand \emph{“intersect”} in current question $q_3$, thus missing \emph{“Lettice”} in the generated self-contained question $\tilde{r}_{3}$ and leading to uncompleted SQL queries in the text-to-SQL stage. In the second case, CQR model misunderstood \emph{“combine”} in the the current question $q_3$, leading to incorrect key word “\texttt{AND}” in the predicted SQL query.

In the lower block of Figure \ref{fig_case_study}, we compare the SQL queries by {\cqrsql} with those by the End-to-End baseline, and visualize the attention patterns of latent variable \texttt{[Z]} on question context. The first case shows the scenario which requires model to inherit history information to resolve the \emph{co-reference} and \emph{ellipsis}. From the heat map, we observe that latent variable \texttt{[Z]} pays more attention to the desirable history information “\emph{volvo}” and “\emph{least accelerate}”. While the second case shows the scenario which requires model to discard confusion history information. In this case, latent variable \texttt{[Z]} pays less attention to the confusion information block “\emph{state of ‘Arizona’}” compared with the desirable ones, which is benefit for correctly SQL parsing. 

\subsection{$\!$\textls[-10]{Transferability~on~Contextual~Text-to-SQL}}
\vspace{-0.1cm}
To verify the transferability of CQR integration on contextual text-to-SQL task, we conduct three \emph{out-of-distribution} experiments as shown in Table \ref{tab_cqr_transferability}. In experiment~\texttt{[\!1\!]}, our {\cqrsql} has better transferability on contextualized questions (${\rm Turn\!\geqslant\!2}$) in \textsc{CoSQL}$_{\rm Dev}$, which contains additional system response and more question turns compared with \textsc{SParC} dataset as in Table \ref{tab_statistic_information}. Beside, in experiment \texttt{[\!2-3\!]}, {\cqrsql} achieves consistently better performances on out-of-distribution contextual questions (${\rm Turn\!\geqslant\!3}$). These results indicate the advantage of {\cqrsql} in robustness contextual understanding for \emph{out-of-distribution} scenarios.

\begin{table}[h]
\centering
\setlength{\abovecaptionskip}{3pt}
\setlength{\belowcaptionskip}{-10pt}
\addtolength{\tabcolsep}{-0.3mm}
\resizebox{0.48\textwidth}{!}{
\begin{tabular}{lcc|c|c|c}  
\toprule[1.0pt]
\hspace{-0.2em}\texttt{\!\![\!\#\!]\!\!\!\!}&\textbf{Train}&\!\textbf{Eval}&\!\textbf{End-to-End}\!&\!\!\textbf{Two-Stage}\!\!&\textbf{{\cqrsql}}\hspace{-0.2em}\\
\midrule[0.3pt]
\hspace{-0.2em}\texttt{\!\![\!1\!]\!\!\!\!}&\textbf{S}$_{\rm Train}^{\rm All}$\!&\!\textbf{C}$_{\rm Dev}^{\rm Turn \geqslant2}$&36.3$_{(+0.0)}$&37.3$_{(+1.0)}$&40.9$_{(+\bm{4.6})}$\hspace{-0.2em}\\
\midrule[0.3pt]
\hspace{-0.2em}\texttt{\!\![\!2\!]\!\!\!\!}&\textbf{S}$_{\rm Train}^{\rm Turn \leqslant2}$\!&\!\textbf{S}$_{\rm Dev}^{\rm Turn \geqslant3}$&41.5$_{(+0.0)}$&42.6$_{(+1.1)}$&45.2$_{(+\bm{3.7})}$\hspace{-0.2em}\\
\midrule[0.3pt]
\hspace{-0.2em}\texttt{\!\![\!3\!]\!\!\!\!}&\textbf{C}$_{\rm Train}^{\rm Turn \leqslant2}$\!&\!\textbf{C}$_{\rm Dev}^{\rm Turn \geqslant3}$&42.4$_{(+0.0)}$&43.2$_{(+0.8)}$&46.7$_{(+\bm{4.3})}$\hspace{-0.2em}\\
\bottomrule[1.0pt]
\end{tabular}
}
\caption{Question match ($\bm{{\rm QM}}$) accuracy results of the \emph{out-of-distribution} experiments. \textbf{S} and \textbf{C} denote \textsc{SParC} and \textsc{CoSQL} datasets respectively.  \textbf{Experiment}~\texttt{\![\!1\!]}:~training models on the training set of \textsc{SParC}$_{{\rm Train}}$, while evaluating them on the questions with context (${\rm Turn\!\geqslant\!2}$) in \textsc{CoSQL}$_{{\rm Dev}}$. \textbf{Experiment} \texttt{\![2-3]}:~training models on the questions at ${\rm Turn\!\leqslant\!2}$, whereas evaluating them on the questions at ${\rm Turn\!\geqslant\!3}$.}
\label{tab_cqr_transferability}
\end{table}

\vspace{-0.2cm}
\section{Related Work}
\vspace{-0.2cm}
\paragraph{Text-to-SQL.}
Spider \cite{yu2018spider} is a well-known cross-domain context-independent text-to-SQL task that has attracted considerable attention. Diverse approaches, such as RAT-SQL \cite{wang2020rat}, \textsc{Bridge} \cite{lin-bridging} and LGESQL \cite{cao2021lge}, have been successful on this task. Recently, with the widespread popularity of dialogue systems and the public availability of \textsc{SParC} \cite{yu2019sparc} and \textsc{CoSQL} \cite{yu2020cosql} datasets, context-dependent text-to-SQL has drawn more attention. \newcite{zhang2019editing} and \newcite{istsql} use previously generated SQL queries to improve the quality of SQL parsing. \newcite{cai-wan-2020-igsql} and \newcite{hui2021dynamic} employ graph neural network to model contextual questions and schema. \newcite{jain-lapata-2021-memory} use a memory matrix to keep track of contextual information. \newcite{chen2021decoupled} decouple context-dependent text-to-SQL task to CQR and context-independent text-to-SQL tasks. Besides, \newcite{yu2020grappa,yu2021score} propose task-adaptive conversational pre-trained model for SQL parsing, and \newcite{scholak2021picard} simply constrain auto-regressive decoders of super-large T5-3B for SQL parsing. In this work, we leverage reformulated self-contained questions in two consistency tasks to enhance contextual dependency understanding for multi-turn text-to-SQL parsing, without suffering from the error propagation of two-stage pipeline methods.\vspace{-0.2cm}

\paragraph{Conversational Question Reformulation (CQR)\!\!} aims to use question context to complete ellipsis and co-references in the current questions. Most works adopt the encoder-decoder architecture with only contextual text as input  \cite{canard,pan2019improving}. 
Besides, CQR are applied to several downstream tasks for enhanced context understanding, such as conversational question answer (CQA) \cite{CQA} and 
conversational passage retrieval (CPR) \cite{DBLP:sigir_2021_cast19}. 
Regrading CQR training for text-to-SQL, We present a recursive CQR method to address long-range dependency and incorporate schema to generate more domain-relevant and semantic-reliable self-contained questions.\vspace{-0.2cm}
\paragraph{Consistency Training.}
To improve model robustness, consistency training \cite{zheng2016improving} has been widely explored in natural language processing by regularizing model predictions to be invariant to small perturbations. The small perturbations can be random or adversarial noise \cite{miyato2018virtual} and data augmentation \cite{zheng2021consistency}. Inspired by consistency training, ExCorD \cite{CQA} trains a classification CQA model that encourage the models to predict similar answers span from the rewritten and original questions.
Different from ExCorD, we combine latent variable with schema grounding consistency task and tree-structured SQL generation consistency task to force model pay more attention to the \emph{co-references} and \emph{ellipsis} in question context. \vspace{-0.1cm}
\section{Conclusions}
\vspace{-0.1cm}
We propose {\cqrsql}, a novel context-dependent text-to-SQL approach that explicitly comprehends the schema and conversational dependency through latent CQR learning. The method introduces a schema enhanced recursive generation mechanism to generate domain-relative self-contained questions, then trains models to map the semantics of self-contained questions and multi-turn question context into the same latent space with schema grounding consistency task and SQL parsing consistency task for adequately context understanding. Experimental results show that {\cqrsql} achieves new state-of-the-art results on two classical context-dependent text-to-SQL datasets \textsc{SParC} and \textsc{CoSQL}.

\section{Limitations}
\vspace{-0.1cm}
Compared to End-to-End approaches as shown in Figure~\ref{fig_preliminary}(a), proposed {\cqrsql} requires more computational cost and GPU memory consumption at each training step, with duel-encoder for question context and self-contained question inputs. Specifically, with batchsize of $32$ and $8$ V100 GPU cards, {\cqrsql} takes $310.7$ seconds to train an epoch on \textsc{SParC} dataset, while End-to-End approache just costs $179.6$ seconds. While compared with previous advanced methods using T5-3B PLM~\cite{scholak2021picard,unifiedskg,rasat}, and multiple task-adaptive PLMs~\cite{hie-sql}, {\cqrsql} is much more computationally efficient. \vspace{-0.1cm}
\section*{Acknowledgments}
We are indebted to the EMNLP2022 reviewers for their detailed and insightful comments on our work. This work was supported in part by the National Natural Science Foundation of China (Grant No. 62276017), and the 2022 Tencent Big Travel Rhino-Bird Special Research Program. \vspace{-5pt}
\bibliography{emnlp_0421}

\begin{thebibliography}{39}
\expandafter\ifx\csname natexlab\endcsname\relax\def\natexlab#1{#1}\fi

\bibitem[{Ba et~al.(2016)Ba, Kiros, and Hinton}]{ln}
Jimmy~Lei Ba, Jamie~Ryan Kiros, and Geoffrey~E Hinton. 2016.
\newblock \href {https://arxiv.org/abs/1607.06450} {Layer normalization}.
\newblock \emph{arXiv preprint arXiv:1607.06450}.

\bibitem[{Cai and Wan(2020)}]{cai-wan-2020-igsql}
Yitao Cai and Xiaojun Wan. 2020.
\newblock \href {https://doi.org/10.18653/v1/2020.emnlp-main.560} {{IGSQL}:
  Database schema interaction graph based neural model for context-dependent
  text-to-{SQL} generation}.
\newblock In \emph{Proceedings of the 2020 Conference on Empirical Methods in
  Natural Language Processing (EMNLP)}, pages 6903--6912, Online. Association
  for Computational Linguistics.

\bibitem[{Cao et~al.(2021)Cao, Chen, Chen, Zhao, Zhu, and Yu}]{cao2021lge}
Ruisheng Cao, Lu~Chen, Zhi Chen, Yanbin Zhao, Su~Zhu, and Kai Yu. 2021.
\newblock \href {https://doi.org/10.18653/v1/2021.acl-long.198} {{LGESQL}: Line
  graph enhanced text-to-{SQL} model with mixed local and non-local relations}.
\newblock In \emph{Proceedings of the 59th Annual Meeting of the Association
  for Computational Linguistics and the 11th International Joint Conference on
  Natural Language Processing (Volume 1: Long Papers)}, pages 2541--2555,
  Online. Association for Computational Linguistics.

\bibitem[{Chen et~al.(2021)Chen, Chen, Li, Cao, Ma, Wu, and
  Yu}]{chen2021decoupled}
Zhi Chen, Lu~Chen, Hanqi Li, Ruisheng Cao, Da~Ma, Mengyue Wu, and Kai Yu. 2021.
\newblock \href {https://doi.org/10.18653/v1/2021.findings-acl.270} {Decoupled
  dialogue modeling and semantic parsing for multi-turn text-to-{SQL}}.
\newblock In \emph{Findings of the Association for Computational Linguistics:
  ACL-IJCNLP 2021}, pages 3063--3074, Online. Association for Computational
  Linguistics.

\bibitem[{Clark et~al.(2020)Clark, Luong, Le, and Manning}]{clark2020electra}
Kevin Clark, Minh-Thang Luong, Quoc~V. Le, and Christopher~D. Manning. 2020.
\newblock \href {https://openreview.net/forum?id=r1xMH1BtvB} {Electra:
  Pre-training text encoders as discriminators rather than generators}.
\newblock In \emph{International Conference on Learning Representations}.

\bibitem[{Dalton et~al.(2020)Dalton, Xiong, Kumar, and
  Callan}]{DBLP:sigir_2021_cast19}
Jeffrey Dalton, Chenyan Xiong, Vaibhav Kumar, and Jamie Callan. 2020.
\newblock \href {https://www.cs.cmu.edu/~callan/Papers/sigir20-dalton.pdf}
  {Cast-19: A dataset for conversational information seeking}.
\newblock In \emph{Proceedings of the 43rd International ACM SIGIR Conference
  on Research and Development in Information Retrieval}, pages 1985--1988.

\bibitem[{Devlin et~al.(2019)Devlin, Chang, Lee, and Toutanova}]{bert}
Jacob Devlin, Ming-Wei Chang, Kenton Lee, and Kristina Toutanova. 2019.
\newblock \href {https://doi.org/10.18653/v1/N19-1423} {{BERT}: Pre-training of
  deep bidirectional transformers for language understanding}.
\newblock In \emph{Proceedings of the 2019 Conference of the North {A}merican
  Chapter of the Association for Computational Linguistics: Human Language
  Technologies, Volume 1 (Long and Short Papers)}, pages 4171--4186,
  Minneapolis, Minnesota. Association for Computational Linguistics.

\bibitem[{Elgohary et~al.(2019)Elgohary, Peskov, and Boyd-Graber}]{canard}
Ahmed Elgohary, Denis Peskov, and Jordan Boyd-Graber. 2019.
\newblock \href {https://doi.org/10.18653/v1/D19-1605} {Can you unpack that?
  learning to rewrite questions-in-context}.
\newblock In \emph{Proceedings of the 2019 Conference on Empirical Methods in
  Natural Language Processing and the 9th International Joint Conference on
  Natural Language Processing (EMNLP-IJCNLP)}, pages 5918--5924, Hong Kong,
  China. Association for Computational Linguistics.

\bibitem[{Hui et~al.(2021)Hui, Geng, Ren, Li, Li, Sun, Huang, Si, Zhu, and
  Zhu}]{hui2021dynamic}
Binyuan Hui, Ruiying Geng, Qiyu Ren, Binhua Li, Yongbin Li, Jian Sun, Fei
  Huang, Luo Si, Pengfei Zhu, and Xiaodan Zhu. 2021.
\newblock \href {https://arxiv.org/abs/2101.01686} {Dynamic hybrid relation
  exploration network for cross-domain context-dependent semantic parsing}.
\newblock In \emph{Proceedings of the AAAI Conference on Artificial
  Intelligence}, volume~35, pages 13116--13124.

\bibitem[{Iyyer et~al.(2017)Iyyer, Yih, and Chang}]{iyyer2017search}
Mohit Iyyer, Wen-tau Yih, and Ming-Wei Chang. 2017.
\newblock \href {https://doi.org/10.18653/v1/P17-1167} {Search-based neural
  structured learning for sequential question answering}.
\newblock In \emph{Proceedings of the 55th Annual Meeting of the Association
  for Computational Linguistics (Volume 1: Long Papers)}, pages 1821--1831,
  Vancouver, Canada. Association for Computational Linguistics.

\bibitem[{Jain and Lapata(2021)}]{jain-lapata-2021-memory}
Parag Jain and Mirella Lapata. 2021.
\newblock \href {https://doi.org/10.1162/tacl_a_00422} {Memory-based semantic
  parsing}.
\newblock \emph{Transactions of the Association for Computational Linguistics},
  9:1197--1212.

\bibitem[{Kim et~al.(2021)Kim, Kim, Park, and Kang}]{CQA}
Gangwoo Kim, Hyunjae Kim, Jungsoo Park, and Jaewoo Kang. 2021.
\newblock \href {https://doi.org/10.18653/v1/2021.acl-long.478} {Learn to
  resolve conversational dependency: A consistency training framework for
  conversational question answering}.
\newblock In \emph{Proceedings of the 59th Annual Meeting of the Association
  for Computational Linguistics and the 11th International Joint Conference on
  Natural Language Processing (Volume 1: Long Papers)}, pages 6130--6141,
  Online. Association for Computational Linguistics.

\bibitem[{Li et~al.(2021)Li, Zhang, Li, Wang, Wu, and Zhang}]{rat-sql-tc}
Yuntao Li, Hanchu Zhang, Yutian Li, Sirui Wang, Wei Wu, and Yan Zhang. 2021.
\newblock \href {https://arxiv.org/pdf/2112.08735.pdf} {Pay more attention to
  history: A context modeling strategy for conversational text-to-sql}.
\newblock \emph{arXiv:2112.08735}.

\bibitem[{Lin et~al.(2020)Lin, Socher, and Xiong}]{lin-bridging}
Xi~Victoria Lin, Richard Socher, and Caiming Xiong. 2020.
\newblock \href {https://doi.org/10.18653/v1/2020.findings-emnlp.438} {Bridging
  textual and tabular data for cross-domain text-to-{SQL} semantic parsing}.
\newblock In \emph{Findings of the Association for Computational Linguistics:
  EMNLP 2020}, pages 4870--4888, Online. Association for Computational
  Linguistics.

\bibitem[{Meng et~al.(2021)Meng, Xiong, Bajaj, tiwary, Bennett, Han, and
  SONG}]{coco}
Yu~Meng, Chenyan Xiong, Payal Bajaj, saurabh tiwary, Paul Bennett, Jiawei Han,
  and XIA SONG. 2021.
\newblock \href
  {https://proceedings.neurips.cc/paper/2021/file/c2c2a04512b35d13102459f8784f1a2d-Paper.pdf}
  {Coco-lm: Correcting and contrasting text sequences for language model
  pretraining}.
\newblock In \emph{Advances in Neural Information Processing Systems},
  volume~34, pages 23102--23114. Curran Associates, Inc.

\bibitem[{Miyato et~al.(2018)Miyato, Maeda, Koyama, and
  Ishii}]{miyato2018virtual}
Takeru Miyato, Shin-ichi Maeda, Masanori Koyama, and Shin Ishii. 2018.
\newblock Virtual adversarial training: a regularization method for supervised
  and semi-supervised learning.
\newblock \emph{IEEE transactions on pattern analysis and machine
  intelligence}, 41(8):1979--1993.

\bibitem[{Pan et~al.(2019)Pan, Bai, Wang, Zhou, and Liu}]{pan2019improving}
Zhufeng Pan, Kun Bai, Yan Wang, Lianqiang Zhou, and Xiaojiang Liu. 2019.
\newblock \href {https://doi.org/10.18653/v1/D19-1191} {Improving open-domain
  dialogue systems via multi-turn incomplete utterance restoration}.
\newblock In \emph{Proceedings of the 2019 Conference on Empirical Methods in
  Natural Language Processing and the 9th International Joint Conference on
  Natural Language Processing (EMNLP-IJCNLP)}, pages 1824--1833, Hong Kong,
  China. Association for Computational Linguistics.

\bibitem[{Qi et~al.(2022)Qi, Tang, He, Wan, Zhou, Wang, Zhang, and Lin}]{rasat}
Jiexing Qi, Jingyao Tang, Ziwei He, Xiangpeng Wan, Chenghu Zhou, Xinbing Wang,
  Quanshi Zhang, and Zhouhan Lin. 2022.
\newblock \href {https://arxiv.org/abs/2205.06983} {Rasat: Integrating
  relational structures into pretrained seq2seq model for text-to-sql}.
\newblock In \emph{arXiv:2205.06983}.

\bibitem[{Qi et~al.(2020)Qi, Yan, Gong, Liu, Duan, Chen, Zhang, and
  Zhou}]{prophetnet}
Weizhen Qi, Yu~Yan, Yeyun Gong, Dayiheng Liu, Nan Duan, Jiusheng Chen, Ruofei
  Zhang, and Ming Zhou. 2020.
\newblock \href {https://doi.org/10.18653/v1/2020.findings-emnlp.217}
  {{P}rophet{N}et: Predicting future n-gram for
  sequence-to-{S}equence{P}re-training}.
\newblock In \emph{Findings of the Association for Computational Linguistics:
  EMNLP 2020}, pages 2401--2410, Online. Association for Computational
  Linguistics.

\bibitem[{Raffel et~al.(2020)Raffel, Shazeer, Roberts, Lee, Narang, Matena,
  Zhou, Li, and Liu}]{t5}
Colin Raffel, Noam Shazeer, Adam Roberts, Katherine Lee, Sharan Narang, Michael
  Matena, Yanqi Zhou, Wei Li, and Peter~J. Liu. 2020.
\newblock \href {http://jmlr.org/papers/v21/20-074.html} {Exploring the limits
  of transfer learning with a unified text-to-text transformer}.
\newblock \emph{Journal of Machine Learning Research}, 21(140):1--67.

\bibitem[{Reimers and Gurevych(2019)}]{sbert}
Nils Reimers and Iryna Gurevych. 2019.
\newblock \href {https://doi.org/10.18653/v1/D19-1410} {Sentence-{BERT}:
  Sentence embeddings using {S}iamese {BERT}-networks}.
\newblock In \emph{Proceedings of the 2019 Conference on Empirical Methods in
  Natural Language Processing and the 9th International Joint Conference on
  Natural Language Processing (EMNLP-IJCNLP)}, pages 3982--3992, Hong Kong,
  China. Association for Computational Linguistics.

\bibitem[{Scholak et~al.(2021)Scholak, Schucher, and
  Bahdanau}]{scholak2021picard}
Torsten Scholak, Nathan Schucher, and Dzmitry Bahdanau. 2021.
\newblock \href {https://doi.org/10.18653/v1/2021.emnlp-main.779} {{PICARD}:
  Parsing incrementally for constrained auto-regressive decoding from language
  models}.
\newblock In \emph{Proceedings of the 2021 Conference on Empirical Methods in
  Natural Language Processing}, pages 9895--9901, Online and Punta Cana,
  Dominican Republic. Association for Computational Linguistics.

\bibitem[{Vaswani et~al.(2017)Vaswani, Shazeer, Parmar, Uszkoreit, Jones,
  Gomez, Kaiser, and Polosukhin}]{transformer}
Ashish Vaswani, Noam Shazeer, Niki Parmar, Jakob Uszkoreit, Llion Jones,
  Aidan~N Gomez, \L~ukasz Kaiser, and Illia Polosukhin. 2017.
\newblock \href
  {https://proceedings.neurips.cc/paper/2017/file/3f5ee243547dee91fbd053c1c4a845aa-Paper.pdf}
  {Attention is all you need}.
\newblock In \emph{Advances in Neural Information Processing Systems},
  volume~30. Curran Associates, Inc.

\bibitem[{Wang et~al.(2020)Wang, Shin, Liu, Polozov, and
  Richardson}]{wang2020rat}
Bailin Wang, Richard Shin, Xiaodong Liu, Oleksandr Polozov, and Matthew
  Richardson. 2020.
\newblock \href {https://doi.org/10.18653/v1/2020.acl-main.677} {{RAT-SQL}:
  Relation-aware schema encoding and linking for text-to-{SQL} parsers}.
\newblock In \emph{Proceedings of the 58th Annual Meeting of the Association
  for Computational Linguistics}, pages 7567--7578, Online. Association for
  Computational Linguistics.

\bibitem[{Wang et~al.(2021)Wang, Ling, Zhou, and Hu}]{istsql}
Run-Ze Wang, Zhen-Hua Ling, Jingbo Zhou, and Yu~Hu. 2021.
\newblock \href {https://arxiv.org/abs/2012.04995} {Tracking interaction states
  for multi-turn text-to-sql semantic parsing}.
\newblock In \emph{Proceedings of the AAAI Conference on Artificial
  Intelligence}.

\bibitem[{Xie et~al.(2022)Xie, Wu, Shi, Zhong, Scholak, Yasunaga, Wu, Zhong,
  Yin, Wang et~al.}]{unifiedskg}
Tianbao Xie, Chen~Henry Wu, Peng Shi, Ruiqi Zhong, Torsten Scholak, Michihiro
  Yasunaga, Chien-Sheng Wu, Ming Zhong, Pengcheng Yin, Sida~I Wang, et~al.
  2022.
\newblock \href {https://arxiv.org/pdf/2201.05966} {Unifiedskg: Unifying and
  multi-tasking structured knowledge grounding with text-to-text language
  models}.
\newblock \emph{arXiv preprint arXiv:2201.05966}.

\bibitem[{Yin and Neubig(2017)}]{decoder}
Pengcheng Yin and Graham Neubig. 2017.
\newblock \href {https://doi.org/10.18653/v1/P17-1041} {A syntactic neural
  model for general-purpose code generation}.
\newblock In \emph{Proceedings of the 55th Annual Meeting of the Association
  for Computational Linguistics (Volume 1: Long Papers)}, pages 440--450,
  Vancouver, Canada. Association for Computational Linguistics.

\bibitem[{Yu et~al.(2021{\natexlab{a}})Yu, Wu, Lin, Wang, Tan, Yang, Radev,
  Socher, and Xiong}]{yu2020grappa}
Tao Yu, Chien-Sheng Wu, Xi~Victoria Lin, Bailin Wang, Yi~Chern Tan, Xinyi Yang,
  Dragomir Radev, Richard Socher, and Caiming Xiong. 2021{\natexlab{a}}.
\newblock \href {https://openreview.net/forum?id=kyaIeYj4zZ} {{G}ra{PP}a:
  grammar-augmented pre-training for table semantic parsing}.
\newblock In \emph{International Conference on Learning Representations}.

\bibitem[{Yu et~al.(2019{\natexlab{a}})Yu, Zhang, Er, Li, Xue, Pang, Lin, Tan,
  Shi, Li, Jiang, Yasunaga, Shim, Chen, Fabbri, Li, Chen, Zhang, Dixit, Zhang,
  Xiong, Socher, Lasecki, and Radev}]{yu2020cosql}
Tao Yu, Rui Zhang, Heyang Er, Suyi Li, Eric Xue, Bo~Pang, Xi~Victoria Lin,
  Yi~Chern Tan, Tianze Shi, Zihan Li, Youxuan Jiang, Michihiro Yasunaga,
  Sungrok Shim, Tao Chen, Alexander Fabbri, Zifan Li, Luyao Chen, Yuwen Zhang,
  Shreya Dixit, Vincent Zhang, Caiming Xiong, Richard Socher, Walter Lasecki,
  and Dragomir Radev. 2019{\natexlab{a}}.
\newblock \href {https://doi.org/10.18653/v1/D19-1204} {{C}o{SQL}: A
  conversational text-to-{SQL} challenge towards cross-domain natural language
  interfaces to databases}.
\newblock In \emph{Proceedings of the 2019 Conference on Empirical Methods in
  Natural Language Processing and the 9th International Joint Conference on
  Natural Language Processing (EMNLP-IJCNLP)}, pages 1962--1979, Hong Kong,
  China. Association for Computational Linguistics.

\bibitem[{Yu et~al.(2021{\natexlab{b}})Yu, Zhang, Polozov, Meek, and
  Awadallah}]{yu2021score}
Tao Yu, Rui Zhang, Alex Polozov, Christopher Meek, and Ahmed~Hassan Awadallah.
  2021{\natexlab{b}}.
\newblock \href {https://openreview.net/forum?id=oyZxhRI2RiE} {Score:
  Pre-training for context representation in conversational semantic parsing}.
\newblock In \emph{International Conference on Learning Representations}.

\bibitem[{Yu et~al.(2018)Yu, Zhang, Yang, Yasunaga, Wang, Li, Ma, Li, Yao,
  Roman, Zhang, and Radev}]{yu2018spider}
Tao Yu, Rui Zhang, Kai Yang, Michihiro Yasunaga, Dongxu Wang, Zifan Li, James
  Ma, Irene Li, Qingning Yao, Shanelle Roman, Zilin Zhang, and Dragomir Radev.
  2018.
\newblock \href {https://doi.org/10.18653/v1/D18-1425} {{S}pider: A large-scale
  human-labeled dataset for complex and cross-domain semantic parsing and
  text-to-{SQL} task}.
\newblock In \emph{Proceedings of the 2018 Conference on Empirical Methods in
  Natural Language Processing}, pages 3911--3921, Brussels, Belgium.
  Association for Computational Linguistics.

\bibitem[{Yu et~al.(2019{\natexlab{b}})Yu, Zhang, Yasunaga, Tan, Lin, Li, Er,
  Li, Pang, Chen, Ji, Dixit, Proctor, Shim, Kraft, Zhang, Xiong, Socher, and
  Radev}]{yu2019sparc}
Tao Yu, Rui Zhang, Michihiro Yasunaga, Yi~Chern Tan, Xi~Victoria Lin, Suyi Li,
  Heyang Er, Irene Li, Bo~Pang, Tao Chen, Emily Ji, Shreya Dixit, David
  Proctor, Sungrok Shim, Jonathan Kraft, Vincent Zhang, Caiming Xiong, Richard
  Socher, and Dragomir Radev. 2019{\natexlab{b}}.
\newblock \href {https://doi.org/10.18653/v1/P19-1443} {{SP}ar{C}: Cross-domain
  semantic parsing in context}.
\newblock In \emph{Proceedings of the 57th Annual Meeting of the Association
  for Computational Linguistics}, pages 4511--4523, Florence, Italy.
  Association for Computational Linguistics.

\bibitem[{Zhang et~al.(2019)Zhang, Yu, Er, Shim, Xue, Lin, Shi, Xiong, Socher,
  and Radev}]{zhang2019editing}
Rui Zhang, Tao Yu, Heyang Er, Sungrok Shim, Eric Xue, Xi~Victoria Lin, Tianze
  Shi, Caiming Xiong, Richard Socher, and Dragomir Radev. 2019.
\newblock \href {https://doi.org/10.18653/v1/D19-1537} {Editing-based {SQL}
  query generation for cross-domain context-dependent questions}.
\newblock In \emph{Proceedings of the 2019 Conference on Empirical Methods in
  Natural Language Processing and the 9th International Joint Conference on
  Natural Language Processing (EMNLP-IJCNLP)}, pages 5338--5349, Hong Kong,
  China. Association for Computational Linguistics.

\bibitem[{Zhao et~al.(2017)Zhao, Zhao, and Eskenazi}]{bow}
Tiancheng Zhao, Ran Zhao, and Maxine Eskenazi. 2017.
\newblock \href {https://doi.org/10.18653/v1/P17-1061} {Learning
  discourse-level diversity for neural dialog models using conditional
  variational autoencoders}.
\newblock In \emph{Proceedings of the 55th Annual Meeting of the Association
  for Computational Linguistics (Volume 1: Long Papers)}, pages 654--664,
  Vancouver, Canada. Association for Computational Linguistics.

\bibitem[{Zheng et~al.(2021)Zheng, Dong, Huang, Wang, Chi, Singhal, Che, Liu,
  Song, and Wei}]{zheng2021consistency}
Bo~Zheng, Li~Dong, Shaohan Huang, Wenhui Wang, Zewen Chi, Saksham Singhal,
  Wanxiang Che, Ting Liu, Xia Song, and Furu Wei. 2021.
\newblock \href {https://doi.org/10.18653/v1/2021.acl-long.264} {Consistency
  regularization for cross-lingual fine-tuning}.
\newblock In \emph{Proceedings of the 59th Annual Meeting of the Association
  for Computational Linguistics and the 11th International Joint Conference on
  Natural Language Processing (Volume 1: Long Papers)}, pages 3403--3417,
  Online. Association for Computational Linguistics.

\bibitem[{Zheng et~al.(2016)Zheng, Song, Leung, and
  Goodfellow}]{zheng2016improving}
Stephan Zheng, Yang Song, Thomas Leung, and Ian Goodfellow. 2016.
\newblock \href {https://arxiv.org/abs/1604.04326} {Improving the robustness of
  deep neural networks via stability training}.
\newblock In \emph{Proceedings of the ieee conference on computer vision and
  pattern recognition}, pages 4480--4488.

\bibitem[{Zheng et~al.(2022)Zheng, Wang, Dong, Wang, and Li}]{hie-sql}
Yanzhao Zheng, Haibin Wang, Baohua Dong, Xingjun Wang, and Changshan Li. 2022.
\newblock \href {https://arxiv.org/abs/2203.07376} {{HIE-SQL}: History
  information enhanced network for context-dependent text-to-sql semantic
  parsing}.
\newblock In \emph{Findings of the Association for Computational Linguistics
  2022}, Online. Association for Computational Linguistics.

\bibitem[{Zhong et~al.(2020)Zhong, Lewis, Wang, and
  Zettlemoyer}]{zhong-etal-2020-grounded}
Victor Zhong, Mike Lewis, Sida~I. Wang, and Luke Zettlemoyer. 2020.
\newblock \href {https://doi.org/10.18653/v1/2020.emnlp-main.558} {Grounded
  adaptation for zero-shot executable semantic parsing}.
\newblock In \emph{Proceedings of the 2020 Conference on Empirical Methods in
  Natural Language Processing (EMNLP)}, pages 6869--6882, Online. Association
  for Computational Linguistics.

\bibitem[{Zhong et~al.(2017)Zhong, Xiong, and Socher}]{zhongSeq2SQL2017}
Victor Zhong, Caiming Xiong, and Richard Socher. 2017.
\newblock \href {https://arxiv.org/abs/1709.00103} {Seq2sql: Generating
  structured queries from natural language using reinforcement learning}.
\newblock \emph{CoRR}, abs/1709.00103.

\end{thebibliography}
\bibliographystyle{acl_natbib}

\clearpage
\appendix
\section{Backbone Architecture}
\label{appendix_backbone}
\subsection{Encoder: Relation-aware Transformer}
Relation-aware Transformer (RAT)~\cite{wang2020rat} is an extension
to Transformer~\cite{transformer} to consider preexisting pairwise relational features between the inputs. For text-to-SQL task, pairwise relational features include the intra-relation of databse schema $D$ and question-schema alignment information. Formally, given input sequence $\bm{x}=\{x_1,x_2,...,x_n\}$, we obtain the initial representations through pre-trained language model $\bm{{\rm H}}^0={\rm PLM}(\bm{x})=\{\bm{{\rm x}}_1,\bm{{\rm x}}_2,...,\bm{{\rm x}}_n\}$. Then $L$ stacked RAT blocks compute the final hidden states $\bm{{\rm H}}^L={\rm RAT}(\bm{x})$ via $\bm{{\rm H}}^l={\rm RAT}_l(\bm{{\rm H}}^{l-1}), l\in [1,L]$ where ${\rm RAT}_l(\cdot)$ is calculated as:
\begin{equation}
\setlength{\abovedisplayskip}{9pt}
\setlength{\belowdisplayskip}{4pt}
\resizebox{.89\linewidth}{!}{$
    \displaystyle
    \begin{aligned}
    \hspace{-0.5em}
    \bm{{\rm Q}}&=\bm{{\rm H}}^{l-1}\bm{{\rm W}}_Q^l,\bm{{\rm K}}=\bm{{\rm H}}^{l-1}\bm{{\rm W}}_K^l,\bm{{\rm V}}=\bm{{\rm H}}^{l-1}\bm{{\rm W}}_V^l \\
    e_{ij}&=\underset{j}{{\rm Softmax}}\left(\frac{\bm{{\rm Q}}_i(\bm{{\rm K}}_j+\bm{{\rm r}}_{ij}^K)^\top}{\sqrt{d_k}}\right)\\[-3mm]
    \bm{{\rm A}}_i\!&=\sum_{j=1}^ne_{ij}\left(\bm{{\rm V}}_j+\bm{{\rm r}}_{ij}^V\right)\\[-1.5mm]
    \tilde{\bm{{\rm A}}}&={\rm LayerNorm}(\bm{{\rm H}}^{l-1}+\bm{{\rm A}})\\
    \bm{{\rm H}}^l\!&={\rm LayerNorm}\left(\tilde{\bm{{\rm A}}}+{\rm FC}({\rm ReLU({\rm FC(\tilde{\bm{{\rm A}}})})})\right).
    \hspace{-2em}
    \end{aligned}$}
\end{equation}
where parameters $\bm{{\rm W}}_Q^l,\bm{{\rm W}}_K^l,\bm{{\rm W}}_V^l\in\mathbb{R}^{d_h\times d_k}$ proj- ect $\bm{{\rm H}}^l$ to queries, keys and values. Embedding $\bm{{\rm r}}_{ij}$ represents the relationship between token $x_i$ and $x_j$. ${\rm LayerNorm}(\cdot)$ is the layer normalization~\cite{ln}, ${\rm FC}(\cdot)$ is the full connected layer.

In {\cqrsql}, the relation type between latent variable \texttt{[Z]} and schema is \textsc{NoMatch}~\cite{wang2020rat}.

\subsection{Decoder: Tree-structured LSTM}
We employ a single layer tree-structured LSTM decoder of \newcite{decoder} to generate the abstract syntax tree (AST) of SQL queries in depth-first, left-to-right order. At each decoding step, the prediction is either 1) \textsc{ApplyRule} action that expands the last non-terminal into a AST grammar rule; or 2) \textsc{SelectColumn} or \textsc{SelectTable} action that chooses a column or table from schema to complete last terminal node. 

Formally, given the final encoder hidden states of ${\rm Question}$, ${\rm Table}$ and ${\rm Column}$ $\bm{{\rm H}}^L=\{\bm{{\rm H}}_q,\bm{{\rm H}}_t,$ $\bm{{\rm H}}_c\}$. The tree-structured decoder is required to generate a sequence of actions $a_t$ to construct the AST which can transfer to standard SQL query $s$, represented as $\mathcal{P}(s|\bm{{\rm H}}^L)=\prod_t\mathcal{P}\!\left(a_t|a_{<t},\bm{{\rm H}}^L\right)$. We adopt a single layer LSTM to produce action sequence, the LSTM states are updated as  $\bm{{\rm c}}_t,\bm{{\rm h}}_t={\rm LSTM}\left([\bm{{\rm a}}_{t-1};\bm{{\rm h}}_{p_t};\bm{{\rm a}}_{p_t};\bm{{\rm n}}_{f_t}],\bm{{\rm c}}_{t-1},\bm{{\rm h}}_{t-1}\right)$ where $[\cdot]$ is the concatenate operation, $\bm{{\rm c}}_t$ is the LSTM cell state, $\bm{{\rm h}}_t$ is the LSTM output hidden state, $\bm{{\rm a}}_{t}$ is the action embedding, $p_t$ is the decoding step of the current parent AST node and $\bm{{\rm n}}_{f_t}$ is the embedding of current node type. We initialize the LSTM hidden state $\bm{{\rm h}}_0$ via attention pooling over the final encoder hidden state $\bm{{\rm H}}^{L}$ as:
\begin{equation}
\setlength{\abovedisplayskip}{5pt}
\setlength{\belowdisplayskip}{5pt}
\resizebox{.67\linewidth}{!}{$
    \displaystyle
    \begin{aligned}
    e_{i}\!&=\underset{i}{{\rm Softmax}}\left(\tilde{\bm{{\rm h}}}_{0}{\rm Tanh}\!\left(\bm{{\rm H}}^{L}_{i}\bm{{\rm W}}_{1}\right)^{\!\top}\right)\\[-3mm]
    \bm{{\rm A}}\!&=\sum_{i=1}^ne_{i}\bm{{\rm H}}^L_i\\[-1.5mm]
    \bm{{\rm h}}_0\!&={\rm Tanh}\left(\bm{{\rm A}}\bm{{\rm W}}_2\right).
    \end{aligned}
$}
\end{equation}
where $\tilde{\bm{{\rm h}}}_0$ is initial attention vector, $ \bm{{\rm W}}_1,\bm{{\rm W}}_2$ are projection parameters. At each decoding step $t$, we employ ${\rm MultiHeadAttention}$~\cite{transformer} on $\bm{{\rm h}}_t$ over $\bm{{\rm H}}^L$ to compute context representation $\bm{{\rm h}}^{\rm ctx}_t$. 

For the prediction of \textsc{ApplyRule} actions, the prediction distribution is computed as:
\begin{equation}
\setlength{\abovedisplayskip}{7pt}
\setlength{\belowdisplayskip}{5pt}
\resizebox{.70\linewidth}{!}{$
    \displaystyle
    \begin{aligned}
    \hspace{-0.5em}
    \mathcal{P}(a_t\!&=\!\textsc{ApplyRule}[R]|a_{<t},\bm{{\rm H}}^L)\\
    &=\underset{R}{{\rm Softmax}}\!\left({\rm MLP}\!_2([\bm{{\rm h}}_t;\bm{{\rm h}}^{\rm ctx}_t])\bm{{\rm W}}_R\right).
    \hspace{-1em}
    \end{aligned}
$}
\end{equation}
where ${\rm MLP}_2$ denotes 2-layer ${\rm MLP}$ with a ${\rm Tanh}$ nonlinearity, $\bm{{\rm W}}_R$ is the classification matrice.

For the prediction of \textsc{SelectTable} actions, the prediction distribution is computed as: 
\begin{equation}
\setlength{\abovedisplayskip}{7pt}
\setlength{\belowdisplayskip}{5pt}
\resizebox{.76\linewidth}{!}{$
    \displaystyle
    \begin{aligned}
    \hspace{-0.5em}
    \mathcal{P}(a_t\!&=\!\textsc{SelectTable}[i]|a_{<t},\bm{{\rm H}}^L)\\
    &=\underset{i}{{\rm Softmax}}\!\left({\rm MLP}\!_2([\bm{{\rm h}}_t;\bm{{\rm h}}^{\rm ctx}_t])\bm{{\rm W}}_{\!t}\bm{{\rm H}}_{t_i}^\top\right).
    \hspace{-1em}
    \end{aligned}
$}
\end{equation}
where $\bm{{\rm H}}_{t_i}$ denotes the encoder hidden state of the $i$-th ${\rm Table}$. The 
prediction of \textsc{SelectColumn} actions is similar to \textsc{SelectTable}.\vspace{-5pt}
\subsection{SQL Parsing Consistency Loss}
\label{appendix_sp}
Given the final encoder hidden states $\bm{{\rm H}}_q^L={\rm RAT}({\rm Seq}\left(\bm{q}_{\leqslant\tau})\right)$ and $\bm{{\rm H}}_r^L={\rm RAT}\left({\rm Seq}(r_{\tau})\right)$ for question context and self-contained question input respectively, the SQL parsing consistency loss $\mathcal{L}^{{\rm SP_{KL}}(t)}$ at each decoding step $t$ is computed as:
\begin{equation}
\setlength{\abovedisplayskip}{7pt}
\setlength{\belowdisplayskip}{7pt}
\resizebox{.86\linewidth}{!}{$
    \displaystyle
    \begin{aligned}
    \hspace{-3em}
    \mathcal{L}^{{\rm SP_{KL}}(t)}&=\mathcal{L}^{{\rm SP_{KL}}(t)}_{\textsc{ApplyRule}}+\mathcal{L}^{{\rm SP_{KL}}(t)}_{\textsc{SelectTable}}\\[1mm]
    &+\mathcal{L}^{{\rm SP_{KL}}(t)}_{\textsc{SelectColumn}}.\\
    \mathcal{L}^{{\rm SP_{KL}}(t)}_{\textsc{ApplyRule}}&={\rm KL}\!\left(\mathcal{P}\!\left(a_t\!=\!\textsc{ApplyRule}|a_{<t},\bm{{\rm H}}_q^L\right)\right.\\
    &\left.\parallel\mathcal{P}\!\left(a_t\!=\!\textsc{ApplyRule}|a_{<t},\bm{{\rm H}}_r^L\right)\right).\\[1mm]
     \mathcal{L}^{{\rm SP_{KL}}(t)}_{\textsc{SelectTable}}&={\rm KL}\!\left(\mathcal{P}\!\left(a_t\!=\!\textsc{SelectTable}|a_{<t},\bm{{\rm H}}_q^L\right)\right.\!\!\!\!\!\!\!\!\!\!\\
    &\left.\parallel\mathcal{P}\!\left(a_t\!=\!\textsc{SelectTable}|a_{<t},\bm{{\rm H}}_r^L\right)\right).
    \hspace{-3em}
    \end{aligned}
$}
\end{equation}
The total SQL parsing consistency loss is computed as $\mathcal{L}^{{\rm SP_{KL}}}=\frac{1}{T}\sum_{t=1}^T\mathcal{L}^{{\rm SP_{KL}}(t)}$, where $T$ denotes the length of action sequence for target SQL AST.

\section{Experiment Details}
\subsection{Detailed Hyperparameters}
We implement {\cqrsql} based on the PyTorch framework\footnote{\url{https://pytorch.org/}} and use 8 Nvidia Tesla V100 32GB GPU cards for all the experiments. Firstly, we trained CQR model with a learning rate of $3e$-$5$ and batch size of $32$. We use the maximum input sequence length as $512$ and the maximum epochs as $25$. We adopt label smooth method with ratio $0.15$ for regularization. During inference for CQR, we set the beam size as 5. 

Regarding training {\cqrsql}, the number of heads is $8$ and hidden size of RAT encoder is $1024$, the dropout rates of encoder and decoder are $0.1$ and $0.2$ respectively. For pre-trained \textsc{Electra}, we adopt layer-wise learning rate decay with coefficient $0.8$ for robust optimization. We train {\cqrsql} on \textsc{SParC} and \textsc{CoSQL} with max training epochs to be $300$ and $350$ respectively.
\subsection{Impact of Weight $\lambda_2$ for Consistency Loss}
\label{appendix_weight}
We vary the weight $\lambda_2$ of consistency loss in $\{1.0,$ $2.0,3.0,4.0\}$ and train {\cqrsql} on \textsc{SParC} and \textsc{CoSQL} datasets, as shown in Table~\ref{tab_cqr_sql_weight}. We observe that \textsc{CoSQL} task desires less CQR knowledge (best choice of $\lambda_2$ is $1.0$) compare with \textsc{SParC} (best choice of $\lambda_2$ is $3.0$), because \textsc{CoSQL} dataset contains much more user focus change questions than \textsc{SParC}, which do not need to be reformulated~\cite{yu2019sparc}.

\begin{table}[h]
\centering
\setlength{\abovecaptionskip}{10pt}
\setlength{\belowcaptionskip}{-5pt}
\addtolength{\tabcolsep}{0.3mm}
\resizebox{0.40\textwidth}{!}{
\begin{tabular}{c|cccc} 
\toprule[1.0pt]\multirow{2}{*}{\!\!\textbf{Weight}~$\lambda_2$}&\multicolumn{2}{c}{\textbf{\textsc{SParC}}$_{\bm{{\rm Dev}}}$}&\multicolumn{2}{c}{\textbf{\textsc{CoSQL}}$_{\bm{{\rm Dev}}}$}\\
&$\bm{{\rm QM}}$&$\bm{{\rm IM}}$&$\bm{{\rm QM}}$&$\bm{{\rm IM}}$\\
\midrule[0.3pt]
\!\!$\lambda_2=1.0$&66.5&47.2&\textbf{58.2}&\textbf{29.4}\\
\!\!$\lambda_2=2.0$&67.1&47.6&58.0&28.3\\
\!\!$\lambda_2=3.0$&\textbf{67.8}&\textbf{48.1}&57.4&27.3\\
\!\!$\lambda_2=4.0$&66.0&46.7&56.7&26.6\\
\bottomrule[1.0pt]
\end{tabular}
}
\caption{Results of {\cqrsql} on \textsc{SParC} and \textsc{CoSQL} datasets with different weights $\lambda_2$ of consistency loss.}
\label{tab_cqr_sql_weight}
\end{table}

\begin{table}[t]
\centering
\setlength{\abovecaptionskip}{5pt}
\setlength{\belowcaptionskip}{10pt}
\addtolength{\tabcolsep}{-1.5mm}
\resizebox{0.485\textwidth}{!}{
\begin{tabular}{l|ccccc}  
\toprule[1.0pt]
\multirow{2}{*}{\textbf{Models}} &\textbf{Turn}~$\bm{1}$& \textbf{Turn}~$\bm{2}$&\textbf{Turn}~$\bm{3}$~~~~&~\textbf{Turn}~$\bm{4}$~&\textbf{Turn}~$\bm{\!>\!4}$ \\
&\#$~293$&\#$~285$&\#$~244$&\#$~114$&\#$~71$ \\
\midrule[0.3pt]
EditSQL$^{\;a}$&50.0&36.7&34.8&43.0&23.9\\
IGSQL$^{\;b}$&53.1&42.6&39.3&43.0&31.0 \\
IST-SQL$^{\;c}$&56.2& 41.0&41.0&41.2&26.8 \\
\textsc{SCoRe}$^{\;d}$&60.8& 53.0&47.5&49.1&32.4\\
\cdashlinelr{1-6}
\textbf{C{\fontsize{9.5pt}{0}\selectfont QR-SQL}}&\textbf{66.2}&\textbf{60.0}&\textbf{54.5}&\textbf{54.4}&\textbf{39.4}\\
\midrule[1.0pt]
\multirow{2}{*}{\textbf{Models}} &$\bm{{\rm Easy}}$& $\bm{{\rm Medium}}$&$\bm{{\rm Hard}}$&$\bm{{\rm Extra}}$& \\
&\#$~483$&\#$~441$&\#$~145$&\#$~134$& \\
\midrule[0.3pt]
EditSQL$^{\;a}$&62.7&29.4&22.8&9.3& \\
IGSQL$^{\;b}$&66.3&35.6&26.4&10.3& \\
IST-SQL$^{\;c}$&66.0&36.2&27.8&10.3&\\
\cdashlinelr{1-6}
\textbf{C{\fontsize{9.5pt}{0}\selectfont QR-SQL}}&\textbf{76.7}&\textbf{55.9}&\textbf{39.9}&\textbf{22.4}&\\
\bottomrule[1.0pt]
\end{tabular}
}
\caption{Detailed $\bm{{\rm QM}}$ results in different interaction turns and goal difficulties on the development set of \textsc{CoSQL} dataset. Detailed results of $^{a\;}$\protect\cite{zhang2019editing},$^{b\;}$\protect\cite{cai-wan-2020-igsql},$^{c\;}$\protect\cite{istsql} and $^{d\;}$\protect\cite{yu2021score} are from the original paper.}
\label{tab_cosql_turn}
\end{table}

\subsection{Detailed Results on \textsc{CoSQL} Task}
As shown in Table~\ref{tab_cosql_turn}, we report the detailed results in different question turns and SQL difficulty levels on the development set of \textsc{CoSQL} dataset. We observe that {\cqrsql} achieves more significant improvement as the interaction turn increases, and consistently outperforms previous works on all SQL difficulty levels.

\subsection{Effects of CQR Integration with Different PLMs}
To further study the effects of CQR integration for contextual text-to-SQL task, we train models in End-to-End, Two-Stage and {\cqrsql} approaches based on different pre-trained language models (PLMs), as shown in Table~\ref{tab_cqr_sql_plm}. We can see that: 1) {\cqrsql} method consistently preforms better than Two-Stage and End-to-End methods, further demonstrating the effectiveness of {\cqrsql} for adequate contextual understanding. 2) \textsc{Coco-LM}~\cite{coco} is superior
to \textsc{Electra}~\cite{clark2020electra} and \textsc{BERT}~\cite{bert}. We argue the reason is that \textsc{Coco-LM} is pre-trained on sequence contrastive
learning with a dual-encoder architecture~\cite{sbert}, which is compatible for our CQR consistency tasks with dual-encoder for question context $\bm{q}_{\leqslant\tau}$ and self-contained question $r_{\tau}$ as inputs.
\begin{table}[hb]
\centering
\setlength{\abovecaptionskip}{5pt}
\setlength{\belowcaptionskip}{-10pt}
\addtolength{\tabcolsep}{-0.25mm}
\resizebox{0.485\textwidth}{!}{
\begin{tabular}{c|c|cccc} 
\toprule[1.0pt]
 \multirow{2}{*}{\bf{PLMs}}&\multirow{2}{*}{\bf{Methods}}&\multicolumn{2}{c}{\textbf{\textsc{SParC}}$_{\bm{{\rm Dev}}}$}&\multicolumn{2}{c}{\textbf{\textsc{CoSQL}}$_{\bm{{\rm Dev}}}$} \\
&&$\bm{{\rm QM}}$&$\bm{{\rm IM}}$&$\bm{{\rm QM}}$&$\bm{{\rm IM}}$ \\
\midrule[0.3pt]
 \multirow{3}{*}{BERT}&End-to-End&58.6&38.2&50.7&20.5 \\
&Two-Stage&60.1&39.3&51.1&22.2 \\
&{\cqrsql}&\textbf{62.5}&\textbf{42.4}&\textbf{53.5}&\textbf{24.6} \\
\midrule[0.3pt]
 \multirow{3}{*}{\textsc{Electra}}&End-to-End&64.9&46.5&56.6&23.9 \\
&Two-Stage&65.8&46.7&56.8&24.6 \\
&{\cqrsql}&\textbf{67.8}&\textbf{48.1}&\textbf{58.2}&\textbf{29.4} \\
\midrule[0.3pt]
 \multirow{3}{*}{\textsc{Coco-LM}}&End-to-End&65.6&45.5&57.1&25.9 \\
&Two-Stage&66.0&46.5&57.8&26.3 \\
&{\cqrsql}&\textbf{68.0}&\textbf{48.8}&\textbf{58.5}&\textbf{31.1} \\
\bottomrule[1.0pt]
\end{tabular}
}
\caption{Results of End-to-End, Two-Stage and {\cqrsql} methods with different PLMs.}
\label{tab_cqr_sql_plm}
\end{table}

\subsection{More Cases}
In this section, we show more cases of error propagation with Two-Stage pipeline method, and {\cqrsql} against End-to-End baseline models.
\begin{figure}[H]
\centering 
\setlength{\abovecaptionskip}{5pt}
\setlength{\belowcaptionskip}{10pt}
\includegraphics[width=7.7cm]{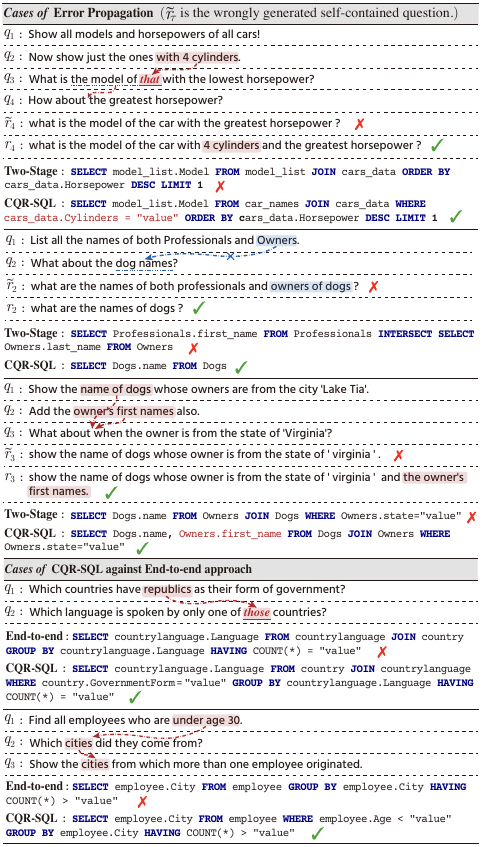}
\caption{Cases on \textsc{SParC} dataset. Upper block shows the cases of error propagation with incorrectly generated self-contained questions $\tilde{r}_{\tau}$ for \textbf{Two-Stage} pipeline methods (as in Figure \ref{fig_preliminary}(c) or \texttt{[7]} in Table \ref{tab_cqr_ablation}). Cases in the lower block show that \textbf{End-to-End} method (as in Figure \ref{fig_preliminary}(a) or \texttt{\![4]\!} in Table \ref{tab_cqr_sql_ablation}) fails to resolve the conversational dependency. }
\label{fig_more_case}
\end{figure}



\end{document}